\documentclass[10pt,twocolumn,letterpaper]{article}

\usepackage[pagenumbers]{cvpr} 

\usepackage[dvipsnames]{xcolor}

\definecolor{cvprblue}{rgb}{0.21,0.49,0.74}
\usepackage[pagebackref,breaklinks,colorlinks,citecolor=cvprblue]{hyperref}

\usepackage{multirow}

\usepackage[accsupp]{axessibility}

\def\paperID{9052} 
\def\confName{CVPR}
\def\confYear{2024}

\title{One-Shot Structure-Aware Stylized Image Synthesis}

\author{Hansam Cho\textsuperscript{\rm 1,2}$^*$,
Jonghyun Lee\textsuperscript{\rm 1,2}$^*$,
Seunggyu Chang\textsuperscript{\rm 1},
Yonghyun Jeong\textsuperscript{\rm 1}$^\dagger$\\
\textsuperscript{\rm 1}NAVER Cloud, 
\textsuperscript{\rm 2}School of Industrial and Management Engineering, Korea University \\
\texttt{\{chosam95, tomtom1103\}@korea.ac.kr}, \\
\texttt{\{seunggyu.chang, yonghyun.jeong\}@navercorp.com}
}

\begin{document}


\newcommand{\xzero}{$\mathbf{x}_0$}
\newcommand{\xt}{$\mathbf{x}_t$}
\newcommand{\xT}{$\mathbf{x}_T$}
\newcommand{\xtzero}{$\mathbf{x}_{\mathbf{t}_0}$}
\newcommand{\xtin}{$\mathbf{x}_{\mathbf{t}_0}^{\mathrm{in}}$}
\newcommand{\xtstyle}{$\mathbf{x}_{\mathbf{t}_0}^{\mathrm{style}}$}
\newcommand{\IstyleA}{$I^{\mathrm{style}}_A$}
\newcommand{\IstyleB}{$I^{\mathrm{style}}_B$}
\newcommand{\IinA}{$I^{\mathrm{in}}_A$}
\newcommand{\IinB}{$I^{\mathrm{in}}_B$}
\newcommand{\DiffA}{$\boldsymbol{\epsilon}^A_\theta$}
\newcommand{\DiffB}{$\boldsymbol{\epsilon}^B_\theta$}
\newcommand{\zsem}{$\mathbf{z}_{\mathrm{sem}}$}
\newcommand{\zsemin}{$\mathbf{z}_{\mathrm{sem}}^{\mathrm{in}}$}
\newcommand{\zsemstyle}{$\mathbf{z}_{\mathrm{sem}}^{\mathrm{style}}$}
\newcommand{\EI}{$E_I$}
\newcommand{\encoder}{$\operatorname{Enc}_\phi$}
\newcommand{\tzero}{$\mathbf{t}_0$}

\newcommand\revision[1]{{\color{blue}#1}}

\twocolumn[{
\renewcommand\twocolumn[1][]{#1}
\maketitle
\vspace{-30pt}
\begin{center}
    \centering
    \captionsetup{type=figure}
    \includegraphics[width=0.90\linewidth]{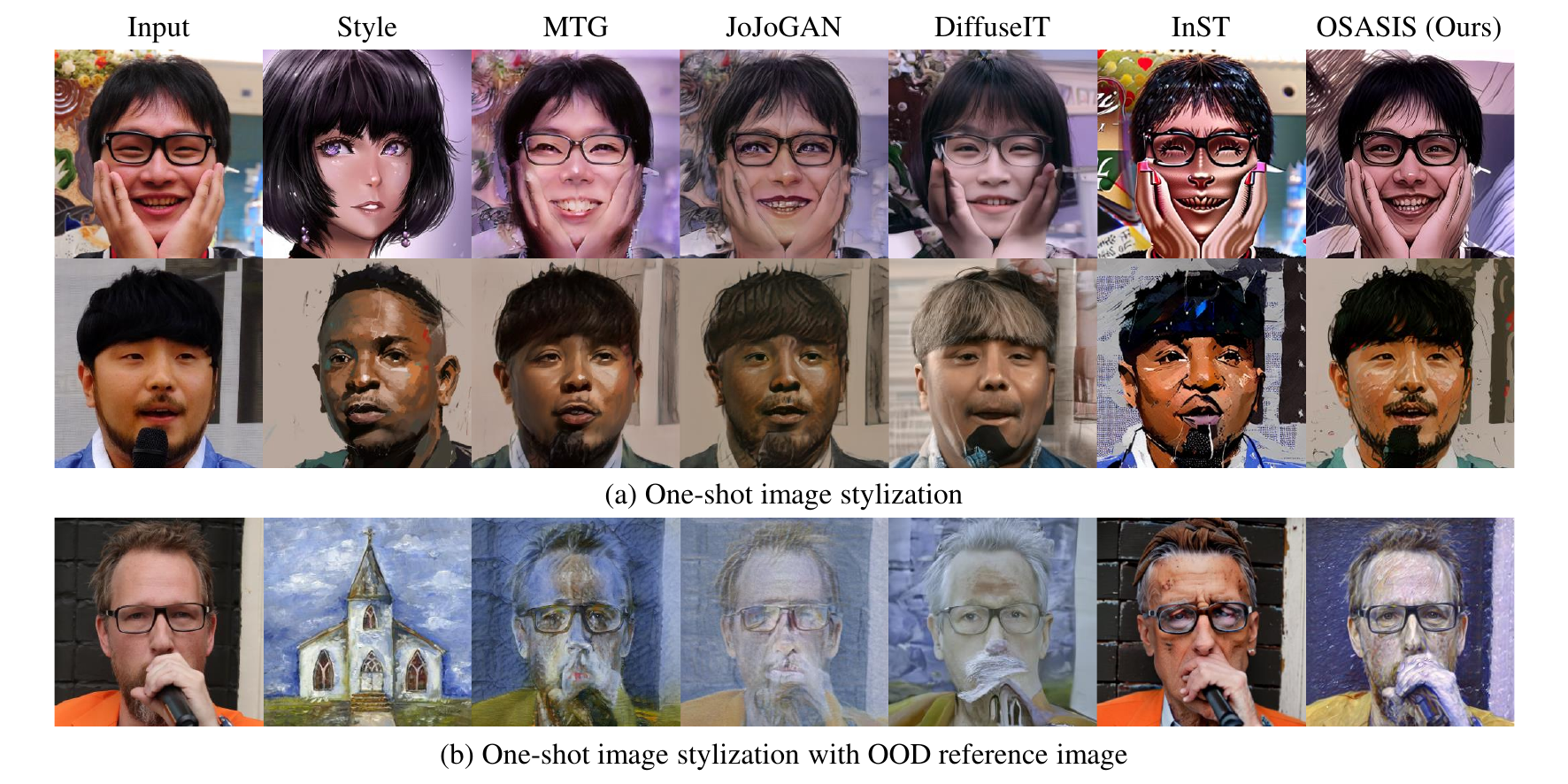}
    \vspace{-15pt}
    \caption{OSASIS is able to (a) stylize an input image with a single reference image while robustly preserving the structure and content of the input image. It is also able to (b) incorporate out-of-domain (OOD) data as the reference image while other baseline methods fail}
    \label{fig:teaser}    
\end{center}
}]

\begin{abstract}
\vspace{-15pt}
While GAN-based models have been successful in image stylization tasks, they often struggle with structure preservation while stylizing a wide range of input images. Recently, diffusion models have been adopted for image stylization but still lack the capability to maintain the original quality of input images. Building on this, we propose OSASIS: a novel one-shot stylization method that is robust in structure preservation. We show that OSASIS is able to effectively disentangle the semantics from the structure of an image, allowing it to control the level of content and style implemented to a given input. We apply OSASIS to various experimental settings, including stylization with out-of-domain reference images and stylization with text-driven manipulation. Results show that OSASIS outperforms other stylization methods, especially for input images that were rarely encountered during training, providing a promising solution to stylization via diffusion models. The source code can be found at \href{https://github.com/hansam95/OSASIS}{https://github.com/hansam95/OSASIS}.
\vspace{-25pt}
\end{abstract}

\def\thefootnote{*}\footnotetext{Work done during an internship at NAVER Cloud.}
\def\thefootnote{$\dagger$}\footnotetext{Corresponding author.}

\section{Introduction}
\vspace{-5pt}
In the literature of generative models, image stylization refers to training a model in order to transfer the style of a reference image to various input images during inference~\cite{ojha2021few, pinkney2020resolution, song2021agilegan}. However, collecting a sufficient number of images that share a particular style for training can be difficult. Consequently, one-shot stylization has emerged as a viable and practical solution, with generative adversarial networks (GANs) showing promising results \cite{chong2022jojogan, kwon2022one, zhang2022towards, zhang2022generalized, zhu2021mind}.

Despite significant advancements in GAN-based stylization techniques, the accurate preservation of an input image's structure continues to pose a significant challenge. This difficulty is particularly pronounced for input images that contain elements infrequently encountered during training, often characterized by complex structural nuances that diverge from those observed in more commonly presented images. Figure~\ref{fig:teaser}(a) illustrates this challenge, where entities such as hands and microphones, when processed through GAN-based stylization, diverge considerably from their original structural integrity. In addition, GAN-based stylization methods often fail to accurately separate the structure and style of the reference image during inference. As shown in Figure~\ref{fig:teaser}(b), the lack of disentanglement results in structural artifacts from reference images bleeding into the stylized image.

Recently, diffusion models (DMs) have shown remarkable performance in various image-related tasks, including high-fidelity image generation \cite{rombach2022high, saharia2022photorealistic}, super resolution \cite{saharia2022image}, and text-driven image manipulation \cite{kim2022diffusionclip, DiffuseIT, Asyrp}. For image stylization, several studies, including DiffuseIT~\cite{DiffuseIT} and InST~\cite{zhang2023inversion}, have been proposed. However, they primarily focus on developing a diffusion model framework tailored to the stylization task. In contrast, our work prioritizes preserving the structure of the input image over solely introducing an appropriate diffusion model for stylization. As illustrated in Figure~\ref{fig:teaser}(a), it can be seen that the capability to preserve structure doesn't stem from  diffusion models itself, but from our methodology.

In this study, we propose \textbf{O}ne-shot \textbf{S}tructure-\textbf{A}ware \textbf{S}tylized \textbf{I}mage \textbf{S}ynthesis (OSASIS), which effectively disentangles the structure and transferable semantics of a style image within the structure of a diffusion model. OSASIS selects an appropriate encoding timestep of a structural latent code to control the strength of structure preservation and enhances its preservation ability through a structure-preserving network. To acquire a semantically meaningful latent, we utilize the semantic encoder proposed in diffusion autoencoders (DiffAE)~\cite{preechakul2022diffusion}. Following the approach of mind the gap (MTG)~\cite{zhu2021mind}, we bridge the domain gap by finetuning a pretrained DM using a combination of directional CLIP losses. Once trained, we find that by properly conditioning the semantic latent code, our method achieves structure-aware image stylization.

We conduct qualitative and quantitative experiments on a wide range of input and style images. By quantitatively extracting data with rare structural elements from the training set (\ie low-density images), we show that OSASIS is robust in structure preservation, outperforming other methods. In addition, we directly optimize the semantic latent code for text-driven manipulation. Combining the optimized latent with the finetuned DM, OSASIS is able to generate stylized images with manipulated attributes.
\section{Background}

\subsection{Diffusion Models}
\vspace{-5pt}
Diffusion models are latent variable models that are trained to reverse a forward process\cite{ho2020denoising}. The forward process, which is defined as a Markov chain with a Gaussian transition defined in Eq.\ref{eq:forward_ddpm}, involves iteratively mapping an image to a predefined prior $\mathcal{N}(\mathbf{0}, \mathbf{I})$ over $T$ steps. DDPM\cite{ho2020denoising} proposes to parameterize the reverse process defined in Eq.~\ref{eq:reverse_ddpm} with a noise prediction network $\epsilon_\theta(\mathbf{x}_t, t)$, which is trained with the loss function $\mathcal{L}_{\textnormal{simple}}$ in Eq.\ref{eq:L_simple}.
\begin{gather}
    \label{eq:forward_ddpm}
    q(\mathbf{x}_t|\mathbf{x}_{t-1}) = N(\mathbf{x}_t ; \sqrt{1-\beta_t}\mathbf{x}_{t-1}, \beta_t \mathbf{I}) \\
    \label{eq:reverse_ddpm}
    p_{\theta}(\mathbf{x}_{t-1}|\mathbf{x}_t) = N(\mathbf{x}_{t-1} ; \mu_{\theta}(\mathbf{x}_t,t), \sigma_t \mathbf{I}) \\
    \quad \text{where} \quad \mu_\theta(\mathbf{x}_t, t) = \frac{1}{\sqrt{1-\beta_t}}(\mathbf{x}_t-\frac{\beta_t}{\sqrt{1-\alpha_t}}\epsilon_\theta(\mathbf{x}_t, t)) \nonumber \\
    \label{eq:L_simple}
    \mathcal{L}_{\textnormal{simple}} = \mathbb{E}_{t,\mathbf{x}_0, \boldsymbol{\epsilon}}\left[\left\|\epsilon_\theta(\mathbf{x}_t, t) - \epsilon)\right\|^2\right], \epsilon \sim \mathcal{N}(\mathbf{0}, \mathbf{I})
\end{gather}

In contrast to DDPM, DDIM \cite{song2020denoising} defines the forward process as non-Markovian and derives the corresponding reverse process as Eq.~\ref{eq:reverse_ddim}, in which $f_\theta(x_t, t)$ is the model's prediction of \xzero\ . DDIM also introduces an image encoding method by deriving ordinary differential equations (ODEs) corresponding with the reverse process. By reversing the ODE, DDIM introduces an image encoding process, defined as Eq.~\ref{eq:forward_ddim}.
\begin{gather}
    \label{eq:forward_ddim}
    \mathbf{x}_{t+1} = \sqrt{\alpha_{t+1}}f_\theta(\mathbf{x}_{t}, t) + \sqrt{1 - \alpha_{t+1}}\epsilon_{\theta}(\mathbf{x}_{t}, t) \\
    \label{eq:reverse_ddim}
    \mathbf{x}_{t-1} = \sqrt{\alpha_{t-1}}f_\theta(\mathbf{x}_{t}, t) +  \sqrt{1 - \alpha_{t-1}}\epsilon_{\theta}(\mathbf{x}_{t}, t) \\
    \quad \text{where} \quad f_\theta(\mathbf{x}_t, t) = \frac{\mathbf{x}_t - \sqrt{1-\alpha_t}\epsilon_\theta(\mathbf{x}_t, t)}{\sqrt{\alpha_t}} \nonumber
\end{gather}

For our work, we adopt specific terminologies to refer to the forward and reverse process of DDPM, in which we call forward DDPM and reverse DDPM, respectively. Similarly, the denoising reverse process of DDIM is referred to as reverse DDIM, while the deterministic image encoding process of DDIM is referred to as forward DDIM.

\begin{figure*}\centering
    \includegraphics[width=0.8\linewidth]{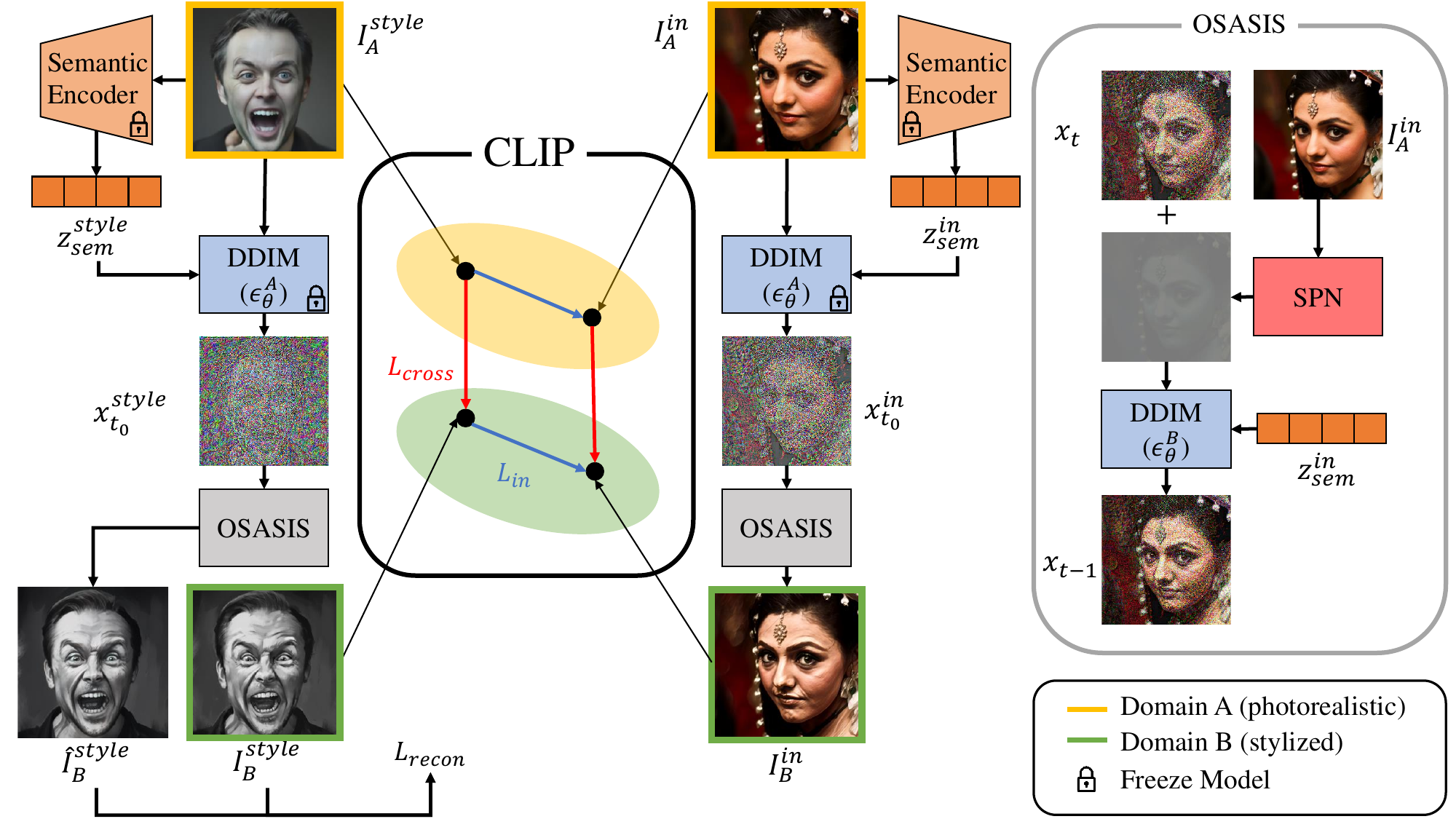}
    \caption{Overview of OSASIS. During finetuning, cross-domain loss compares the photorealistic image (bounded yellow) to its stylized counterparts (bounded green). 
    Concurrently, the in-domain loss gauges the alignment of directional shifts within the same domain, which are delineated by yellow and green. Reconstruction loss compares the original style image with a reconstructed counterpart. Intuitively, the combination of the directional losses guarantees that for each iteration the generated \IinB\ is positioned for projection vectors from \IstyleB\ and \IinA\ to be collinear to its cross-domain and in-domain counterparts in the CLIP space.}
    \vspace{-10pt}
    \label{fig:method}
\end{figure*}

\subsection{Diffusion Autoencoders}
\vspace{-5pt}
DiffAE \cite{preechakul2022diffusion} proposes a semantic encoder \encoder\ that encodes a given image \xzero\ to a semantically rich latent variable \zsem, represented as:
\vspace{-5pt}
\begin{gather}
    \label{eq:semantic_encoder}    \mathbf{z}_{\mathrm{sem}}=\operatorname{Enc}_\phi\left(\mathbf{x}_0\right).
\end{gather}
This latent variable has been shown to be linear, decodable, and semantically meaningful, hence being an attractive property that our model seeks to leverage. Similar to the aforementioned forward DDIM, DiffAE is also able to encode an image given \zsem\ to a fully reconstructable latent \xT\ by Eq.~\ref{eq:forward_diffae}, in which we denote forward DiffAE. Correspondingly, the forward DiffAE encoded latent \xT\ can be decoded by conditioning itself and \zsem\ to the reverse process defined as Eq.~\ref{eq:reverse_diffae}, referred to as reverse DiffAE.
\begin{gather}
    \label{eq:forward_diffae}
    \mathbf{x}_{t+1} = \sqrt{\alpha_{t+1}}f_\theta(\mathbf{x}_{t}, t, \mathbf{z}_{\mathrm{sem}}) + \sqrt{1 - \alpha_{t+1}}\epsilon_{\theta}(\mathbf{x}_{t}, t, \mathbf{z}_{\mathrm{sem}}) \\
    \label{eq:reverse_diffae}
    \mathbf{x}_{t-1} = \sqrt{\alpha_{t-1}}f_\theta(\mathbf{x}_{t}, t, \mathbf{z}_{\mathrm{sem}}) +  \sqrt{1 - \alpha_{t-1}}\epsilon_{\theta}(\mathbf{x}_{t}, t, \mathbf{z}_{\mathrm{sem}}) \\
    \quad \text{where} \quad f_\theta(\mathbf{x}_t, t, \mathbf{z}_{\mathrm{sem}}) = \frac{\mathbf{x}_t - \sqrt{1-\alpha_t}\epsilon_\theta(\mathbf{x}_t, t, \mathbf{z}_{\mathrm{sem}})}{\sqrt{\alpha_t}} \nonumber
\end{gather}

\section{Methods}
\vspace{-5pt}
Our approach aims to achieve effective stylization by initially disentangling the structural and semantic information of images. We define structural information as the overall outline of an image, whereas we further deconstruct the semantics of an image into a combination of content and style. To initially disentangle the semantics from the structure, we employ two distinct latent codes: the structural latent code \xtzero\ and the semantic latent code \zsem. We finetune a pretrained DDIM \DiffA\ conditioned on the semantic latent code \zsem\ via CLIP directional losses in order to bridge the domain gap between the input and style images.  Once finetuned, we control the amount of low-level visual features (\eg texture and color), referred to as style, and high-level visual features (\eg object and identity), referred to as content during inference. Proper conditioning of \zsem\ to the finetuned DDIM allows us to achieve this control, effectively performing stylization. Furthermore, we directly optimize the semantic latent code \zsem\ for text-driven manipulation. By combining the optimized latent with the finetuned DDIM, OSASIS is able to produce stylized images with manipulated attributes. Figure~\ref{fig:method} provides an overview of our method.
\vspace{-5pt}

\subsection{Training}
\vspace{-5pt}
To ensure that the changes in the CLIP embedding space occur in the desired direction, we prepare a single image \IstyleB\ from a stylized domain (denoted domain B) and aim to convert it to a photorealistic domain (denoted domain A). Recent studies have shown that pretrained DMs can generate domain-specific images based on unseen domain images~\cite{kim2022diffusionclip, meng2022sdedit}. Building on this, we utilize a pretrained DDPM $\epsilon_\theta$ to generate a single image \IstyleA\ from domain A that is semantically aligned with \IstyleB. \IstyleB\ is encoded to a specific timestep \tzero\ by utilizing the forward DDPM:
\begin{gather}
    \label{eq:forward_ddpm2}
    \mathbf{x}_{\mathbf{t}_0} = \sqrt{\alpha_{\mathbf{t}_0}}\mathbf{x}_0 + \sqrt{1-\alpha_{\mathbf{t}_0}}\mathbf{z}, \quad \mathbf{z} \sim \mathcal{N}(\mathbf{0}, \mathbf{I}),
\end{gather}
and subsequently from $\mathbf{x}_{\mathbf{t}_0}$, \IstyleA\ is generated by following the reverse DDPM:
\begin{gather}
    \label{eq:reverse_ddpm2}
     \mathbf{x}_{t-1} = \frac{1}{\sqrt{1-\beta_t}}(\mathbf{x}_t-\frac{\beta_t}{\sqrt{1-\alpha_t}}\epsilon_\theta(\mathbf{x}_t, t)) + \sigma_t\mathbf{z},
\end{gather}
where $\mathbf{z} \sim \mathcal{N}(\mathbf{0}, \mathbf{I})$. After initializing \IstyleA\ and \IstyleB, we proceed to freeze a pretrained DDIM \DiffA\ and the semantic encoder \encoder\ proposed by DiffAE, which is utilized during the image encoding process. Additionally, we create a copy of \DiffA\ called \DiffB, which is finetuned through a combination of CLIP directional losses and a reconstruction loss. During training, we note that \IinA\ is generated from the pretrained DiffAE. Consequently, our method enables training without the necessity for a dataset.
\vspace{-10pt}

\paragraph{Structural Latent Code}
\DiffB\ optimizes towards generating \IinB\ that reflects the semantics of the style image. However, since \IstyleA\ and \IstyleB\ stays fixed while \IinA\ is constantly generated, it is crucial to carefully choose an encoding process that generates \IinB\ that preserves the structural integrity of \IinA. In order to achieve this, we utilize forward DiffAE to encode \IinA\ by computing Eq.~\ref{eq:forward_diffae}. First, \IinA\ is encoded into a semantic latent code \zsemin\ = \encoder(\IinA). By following the forward process, \zsemin\ is conditioned to the frozen DDIM \DiffA,
\begin{gather}
    \label{eq:forward_diffae_in}
    \mathbf{x}_{t+1} = \sqrt{\alpha_{t+1}}f_\theta(\mathbf{x}_{t}, t, \mathbf{z}_{\mathrm{sem}}^{\mathrm{in}}) + \sqrt{1 - \alpha_{t+1}}\epsilon^A_{\theta}(\mathbf{x}_{t}, t, \mathbf{z}_{\mathrm{sem}}^{\mathrm{in}})
\end{gather}
resulting in \IinA\ encoded to a structural latent code \xtin. The input image is encoded to a specific timestep \tzero\, which can be adjusted to control the level of structure preservation. 
\vspace{-10pt}

\paragraph{Structure-Preserving Network}
Although the structural latent code \xtin\ succeeds in preserving the overall structure of the generated images \IinA, the encoding process defined in Eq.~\ref{eq:forward_diffae_in} inherently adds noise, which inevitably results in the loss of structural information. To address this, we introduce a structure-preserving network (SPN), which utilizes a 1x1 convolution that effectively preserves the spatial information and structural integrity of \IinA. To generate the output of the next timestep $\mathbf{x}_{t-1}$, we use reverse DiffAE with SPN:
\vspace{-15pt}
\begin{gather}
    \label{eq:output_SPN}
    \mathbf{x}_{t}^{SPN} = SPN(I^\mathrm{in}_A) \\
    \label{eq:combine_SPN}
    \mathbf{x}'_{t} = \mathbf{x}_{t} + \lambda_{SPN} * \mathbf{x}_{t}^{SPN} \\
    \label{eq:reverse_diffae_in}
    \mathbf{x}_{t-1} = \sqrt{\alpha_{t-1}}f_\theta(\mathbf{x}'_{t}, t, \mathbf{z}_{\mathrm{sem}}^{\mathrm{in}}) + \sqrt{1 - \alpha_{t-1}}\epsilon^B_{\theta}(\mathbf{x}_{t}', t, \mathbf{z}_{\mathrm{sem}}^{\mathrm{in}})
\end{gather}
We add the output of the SPN to $\mathbf{x}_{t}$, \eg the previous timestep output of \DiffB, and feed it into our training target \DiffB. We regulate the degree of spatial information reflected in the model by multiplying the output of the SPN $\mathbf{x}_{t}^{SPN}$ by $\lambda_{SPN}$. After fully reversing the timesteps, \IinB\ is generated.
\vspace{-20pt}

\paragraph{Loss Function}
Inspired by MTG \cite{zhu2021mind}, we train \DiffB\ by optimizing our total loss, which is comprised of the cross-domain loss, in-domain loss, and reconstruction loss. The cross-domain loss aims to align the direction of changes from domain A to domain B, ensuring that the change from \IinA\ to \IinB\ is kept consistent with the change from \IstyleA\ to \IstyleB. Although the cross-domain loss provides the changes in semantic information for the model to optimize upon, it often leads to unintended changes when implemented alone. Hence the in-domain loss is introduced to provide additional information, measuring the similarity of changes within both domains A and B. 

The reconstruction loss provides additional guidance in capturing the cross-domain change from \IstyleA\ to \IstyleB. Similar to our encoding process of Eq.~\ref{eq:forward_diffae_in}, \IstyleA\ is encoded to a structural latent code \xtstyle\ conditioned on semantic latent code \zsemstyle. Following Eq.~\ref{eq:output_SPN}-~\ref{eq:reverse_diffae_in}, \ie the process of generating the output of the next timestep conditioned on \zsemstyle, $\hat{I}^{\mathrm{style}}_B$ is generated. The reconstruction loss is calculated by comparing $\hat{I}^{\mathrm{style}}_B$ with \IstyleB, comprised of the $L_1$ loss, perceptual similarity loss~\cite{zhang2018unreasonable}, and the $L_1$ CLIP embedding loss. Detailed information regarding the loss function and experimental setup is available in the supplementary material Section~\ref{supp:loss}.

\vspace{-5pt}
\subsection{Sampling}
\vspace{-5pt}
\paragraph{Mixing Content and Style}
Once trained, the model \DiffB\ is capable of stylizing images from domain A to B. Stylizing an image involves mixing two images in its latent space. While this process is straightforward with StyleGAN~\cite{karras2019style}, it is still an ongoing research area for diffusion models. Unlike the original DiffAE which conditions a single semantic latent code \zsem\ to the feature maps of DDIM, we discover that properly conditioning \zsem\ to the feature maps of \DiffB\ achieves content and style mixing. This is done by conditioning \zsemstyle\ to its low-level feature maps to transfer the style of a style image, and conditioning \zsemin\ to its high-level feature maps to transfer the content of an input image. The change point of conditioning is set as $f_{ch}$. Since \DiffB\ is a UNet-based model, conditioning the two latents is symmetrical. To preserve the structural information of the input image, we use \xtin\ as the structural latent code. The detailed process of sampling is described in the supplementary material Secition~\ref{supp:mix_cont_style}.
\vspace{-10pt}

\paragraph{Text-driven Manipulation}
Instead of optimizing a model, we directly optimize the semantic latent code of input image \zsemin\ to achieve text-driven manipulation. Similar to previous works~\cite{kim2022diffusionclip}, we use CLIP directional loss for optimization. After optimization, the optimized \zsemin\ can be passed into the \DiffB to incorporate the style into the image. Comprehensive details about the experimental approach for text-driven manipulation are available in the supplementary material Section~\ref{supp:text_mani}.
\vspace{-5pt}
\section{Experiments}
\begin{figure}[t]
    \centering
    \includegraphics[width=\linewidth]{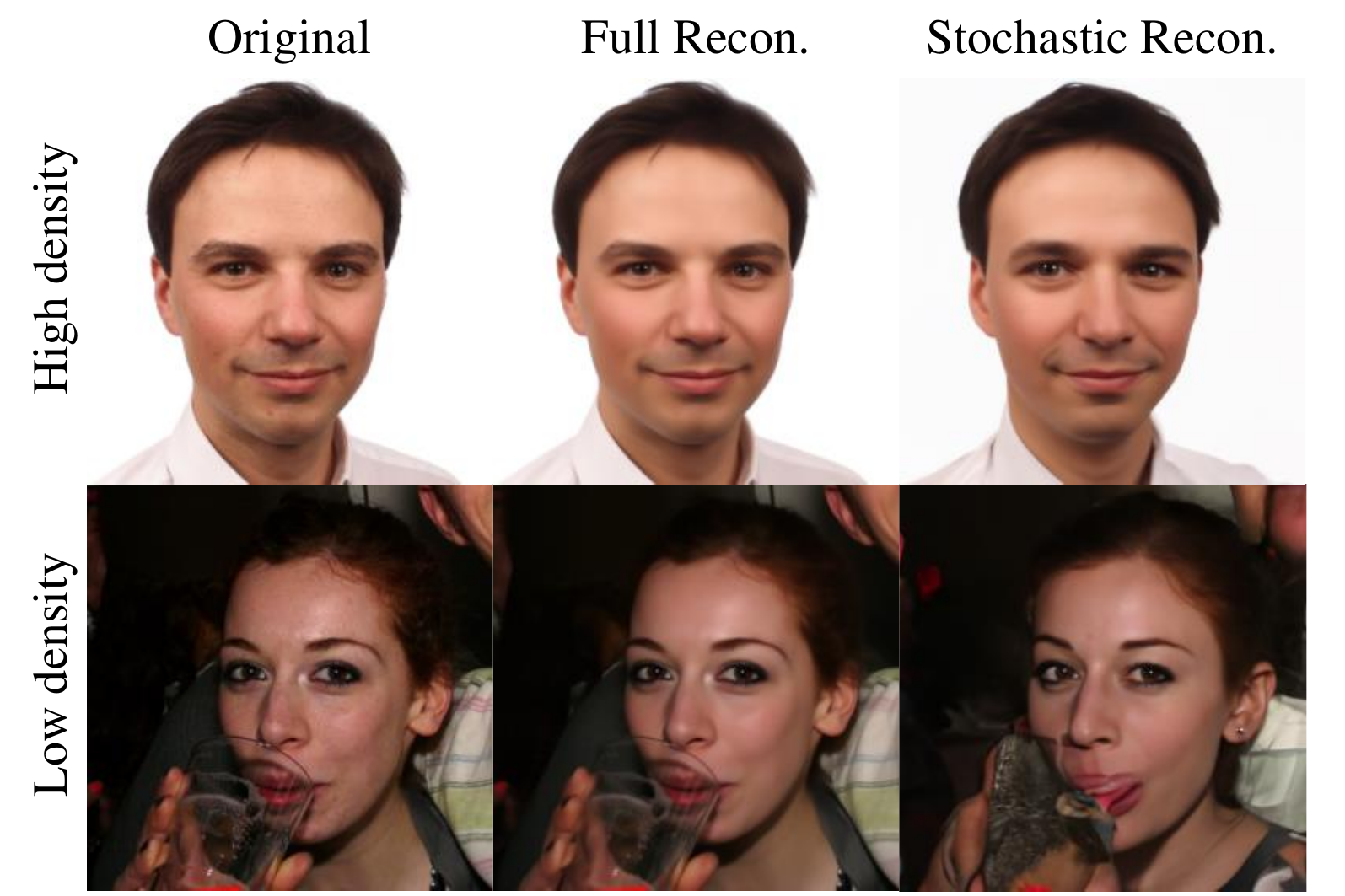}
    \caption{High and Low-density images. Full Recon. refers to reconstruction via conditioning its encoded \zsem\ and \xT, whereas Stochastic Recon. refers to reconstruction via conditioning its encoded \zsem\ and \xT\ $\sim \mathcal{N}(\mathbf{0}, \mathbf{I})$.}
    \vspace{-15pt}
    \label{fig:recon_comparison}
\end{figure}

\vspace{-5pt}
For evaluating our approach, we focus on images from low-density regions that include rarely encountered attributes during training. This is ideal for demonstrating the structure-preserving ability of our method as low-density region images contain diverse objects and occlusions that obscure the subject. To select these images, we leverage the property that the encoded stochastic subcode \xT\ tends to show residuals of the original input image rather than being normally distributed, as shown by DiffAE~\cite{preechakul2022diffusion}. We randomly select 20,000 images from the FFHQ dataset~\cite{karras2019style}, which is the dataset \DiffA\ was trained on. We reconstruct each image using its semantic subcode \zsem\ and stochastic subcode $\mathbf{x}_T \sim\ \mathcal{N}(\mathbf{0}, \mathbf{I})$ (\ie stochastic reconstruction). We compare the reconstructed image with the original image using perceptual similarity loss~\cite{zhang2018unreasonable} to determine the quality of the reconstruction. We hypothesize that high-density region images that contain frequently encountered attributes would be reconstructed accurately, whereas those from low-density regions would not. Figure~\ref{fig:recon_comparison} shows that our hypothesis is well supported. Finally, we select the images from the top 100 highest LPIPS score group (\ie low-density) and lowest LPIPS score group (\ie high-density).

\begin{figure*}\centering
    \includegraphics[width=\linewidth]{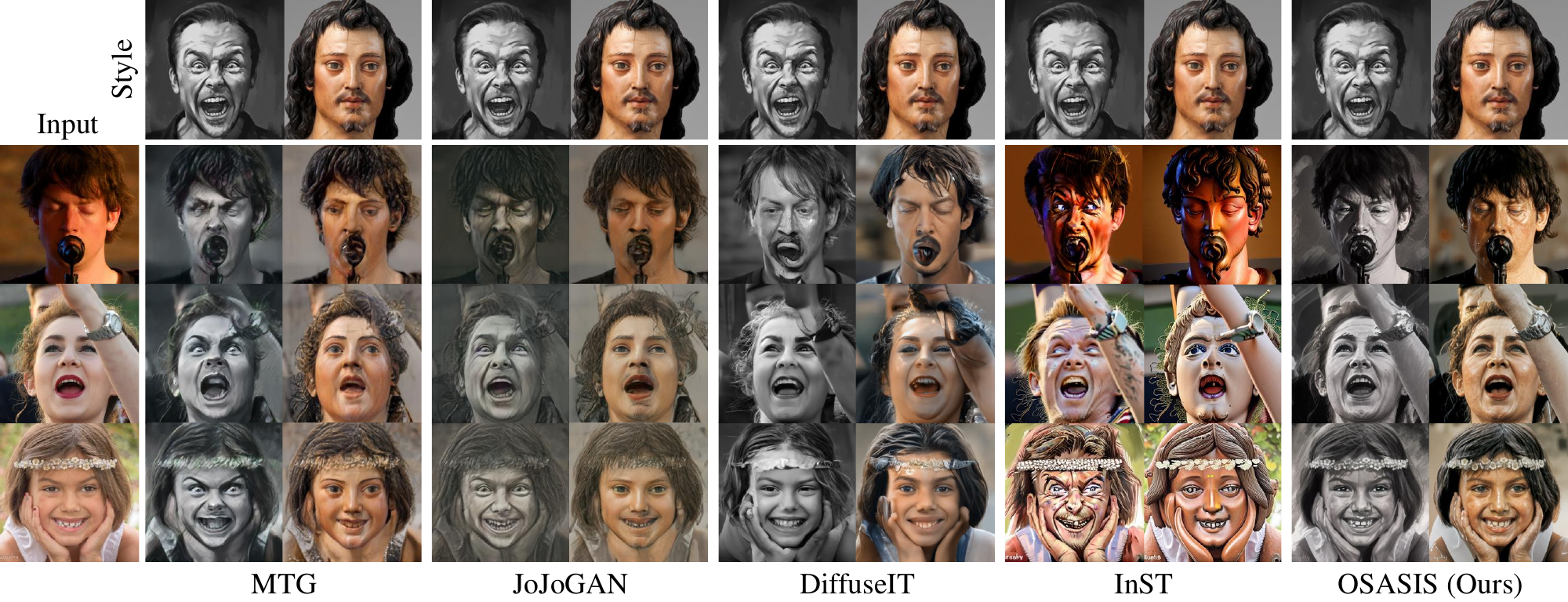}
    \vspace{-20pt}
    \caption{Comparison with other stylization methods. Note that our method successfully preserves the low-density attributes while other baseline methods fail to do so.}
    \vspace{-10pt}
    \label{fig:comparison}
\end{figure*}

\begin{figure*}\centering
    \includegraphics[width=\linewidth]{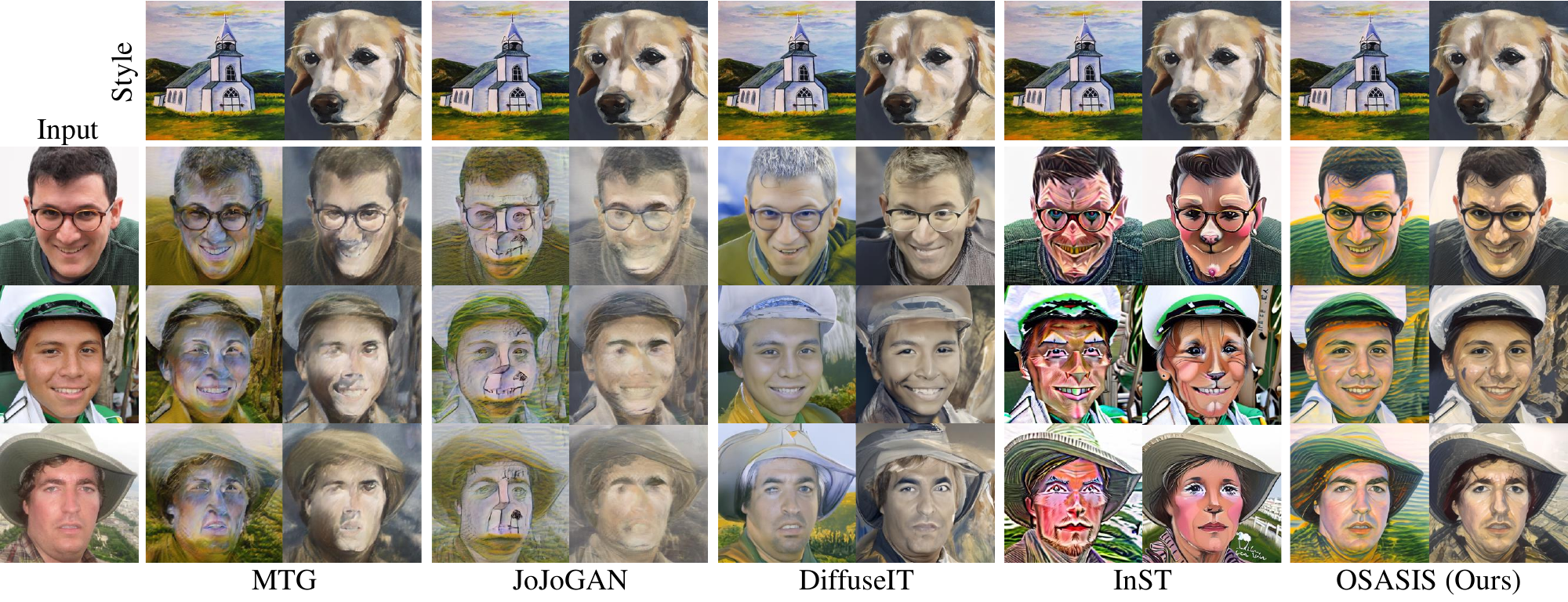}
    \vspace{-20pt}
    \caption{Stylization with OOD reference images. Due to the limited capabilities of GAN-based inversion methods, the baseline methods fail in disentangling the structure and semantics of the style image. This results in structural artifacts being transferred into the output image, whereas OSASIS successfully extracts only the semantics.}
    \vspace{-15pt}
    \label{fig:OOD style}
\end{figure*}

\subsection{Qualitative Comparison}
\vspace{-5pt}
In Figure~\ref{fig:comparison}, we present a comparative analysis of the performance of OSASIS against other stylization methods. Our results demonstrate that OSASIS outperforms other methods in terms of preserving the overall structure while stylizing. MTG~\cite{zhu2021mind} and JoJoGAN~\cite{chong2022jojogan} use outdated inversion methods that struggle to preserve the diverse structure of input images. Nonetheless, recent advancements in GAN-based inversion techniques have demonstrated significant improvements in handling out-of-distribution input images for editing purposes~\cite{roich2022pivotal, wang2022high, xu2023n}. To ensure a fair comparison, we employ HFGI~\cite{wang2022high} for MTG and JoJoGAN. Despite these adjustments, OSASIS remains distinguished in its structural preservation capabilities. Furthermore, GAN-based methods produce unintended modifications, such as changes in facial expressions. Since DiffuseIT~\cite{DiffuseIT} stylizes images without training, they struggle to overcome the domain gap between the input and style images. InST~\cite{zhang2023inversion} utilizes textual inversion to deduce the style image's concept and subsequently conditions its generation procedure on this concept. However, the guidance strategy outlined in~\cite{ho2021classifier} tends to produce style-concentrated images, leading to unintended variations such as changes in facial expressions and identities. Moreover, as noted in InST, there are difficulties in faithfully transferring the color of the style images. More qualitative comparison results are provided in the supplementary material Figure~\ref{fig:supp_face}.

\begin{table}
    \centering
    \begin{tabular}{l|ccc}
    \hline
    \multirow{2}{*}{Methods} & \multicolumn{3}{c}{ArtFID$\downarrow$} \\
    & AAHQ & MetFaces & Prev \\
    \hline
    MTG+HFGI & 36.39 & \textbf{38.02} & 37.27 \\
    JoJoGAN+HFGI & 40.41 & 44.74 & 41.09 \\
    DiffuseIT & 44.93 & 53.35 & 48.18 \\
    InST & 38.16 & 50.33 & 35.86 \\
    \textbf{OSASIS(Ours)} & \textbf{34.89} & 43.20 & \textbf{33.20} \\
    \hline
    \multirow{2}{*}{Methods} & \multicolumn{3}{c}{ID Similarity$\uparrow$} \\
    & AAHQ & MetFaces & Prev \\
    \hline
    MTG+HFGI & 0.3730 & 0.4656 & 0.4063 \\
    JoJoGAN+HFGI & 0.5145 & 0.5207 & 0.4743 \\
    DiffuseIT & \textbf{0.6992} & 0.7158 & 0.6994\\
    InST & 0.2253 & 0.2188 & 0.2238 \\
    \textbf{OSASIS(Ours)} & 0.6825 & \textbf{0.7323} & \textbf{0.7029} \\
    \hline
    \multirow{2}{*}{Methods} & \multicolumn{3}{c}{Structure Distance$\downarrow$} \\
    & AAHQ & MetFaces & Prev \\
    \hline
    MTG+HFGI & 0.0386 & 0.0350 & 0.0360 \\
    JoJoGAN+HFGI & 0.0411 & 0.0454 & 0.0430 \\
    DiffuseIT & \textbf{0.0309} & 0.0300 & \textbf{0.0310} \\
    InST & 0.0492 & 0.0443 & 0.0488 \\
    \textbf{OSASIS(Ours)} & 0.0361 & \textbf{0.0295} & 0.0391 \\
    \hline
    \end{tabular}
    \vspace{-5pt}
    \caption{Quantitative comparison. ArtFID evaluates the pertinence of stylization, whereas ID similarity and structure distance measure whether the stylized image stays true to its original input.}
    \vspace{-15pt}
    \label{table:quant_compare}
\end{table}

\subsection{Quantitative Comparison}
\vspace{-5pt}
We conduct a quantitative comparison with other methods.
By using ArtFID~\cite{wright2022artfid} as a metric for effective stylization and the identity similarity measure with ArcFace~\cite{deng2019arcface} to assess content preservation. For measuring structure preservation, we employ the structure distance metric~\cite{tumanyan2022splicing}. For our source of style images, we select five style images from each of three datasets: \romannumeral1) AAHQ~\cite{liu2021blendgan}, \romannumeral2) MetFaces ~\cite{karras2020training}, \romannumeral3) style images used in previous researches. In Table~\ref{table:quant_compare}, we evaluate OSASIS against MTG~\cite{zhu2021mind}, JoJoGAN~\cite{chong2022jojogan}, DiffuseIT~\cite{DiffuseIT}, and InST~\cite{zhang2023inversion}. For our initial pretrained \DiffA\ and all baseline models, we use publicly available pretrained models that were trained on FFHQ. As previously mentioned, HFGI~\cite{wang2022high} is employed to invert input images for MTG and JoJoGAN. Note that Table~\ref{table:quant_compare} presents outcomes only for low-density images, while a comprehensive comparison is presented in the supplementary material Table~\ref{table:quant_compare_full}. Our results indicate that while MTG occasionally achieves a better ArtFID score than our model, we outperform significantly in terms of identity similarity and structure preservation. Additionally, while DiffuseIT excels at preserving the identity and the structure of the image, it exhibits inferior stylization results compared to GAN-based methods due to a domain gap between the input and style images.

\begin{figure*}[h]\centering
    \includegraphics[width=\linewidth]{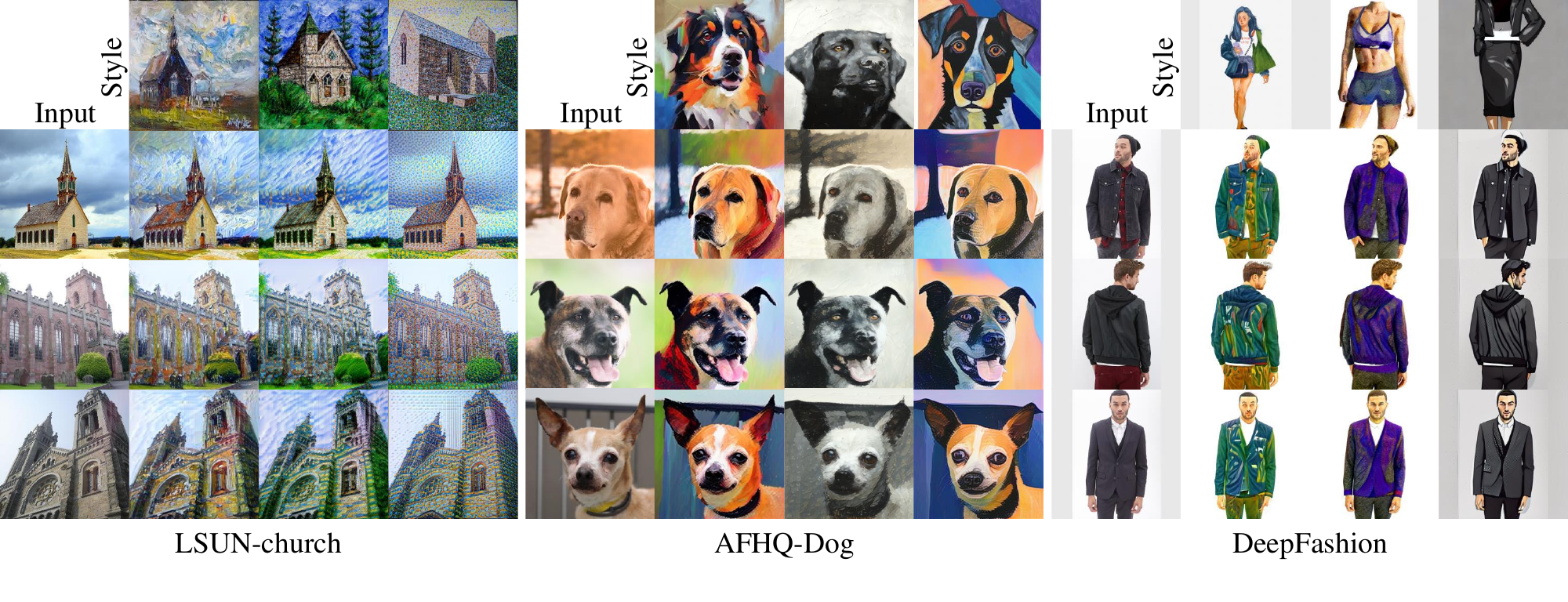}
    \vspace{-30pt}
    \caption{Stylization result of OSASIS on LSUN-church, AFHQ-dog, and DeepFashion.}
    \vspace{-10pt}
    \label{fig:supp_other}
\end{figure*}

\subsection{OOD Reference Image}
\vspace{-5pt}
OSASIS is able to stylize images with out-of-domain (OOD) reference images, a feature that is not commonly seen in GAN-based methods. OSASIS can effectively disentangle semantics from the structure, resulting in only the style factor of the style image being transferred to the input image. In contrast, GAN-based methods have entangled style codes in terms of structure and semantics, which makes it difficult to transfer only the style factor from the reference image. As shown in Figure~\ref{fig:OOD style}, OSASIS is able to stylize images with OOD reference images while preserving its content, whereas other baseline methods suffer from severe artifacts. Although DiffuseIT and InST manage to avoid severe artifacts, they still struggle with addressing domain gap and handling strong concept conditioning.

\subsection{Stylization Results on Other Datasets}
\vspace{-5pt}
To evaluate the generalization capabilities of our method across various datasets, we executed stylization on several distinct datasets: AFHQ-dog \cite{choi2020stargan}, LSUN-church \cite{yu2015lsun}, and DeepFashion \cite{liuLQWTcvpr16DeepFashion}. Figure~\ref{fig:supp_other} displays the efficacy and versatility of our approach across these diverse datasets. Notably, our method exhibits proficiency beyond facial stylization, adeptly adapting to various image domains.

\begin{figure}\centering
    \includegraphics[width=\linewidth]{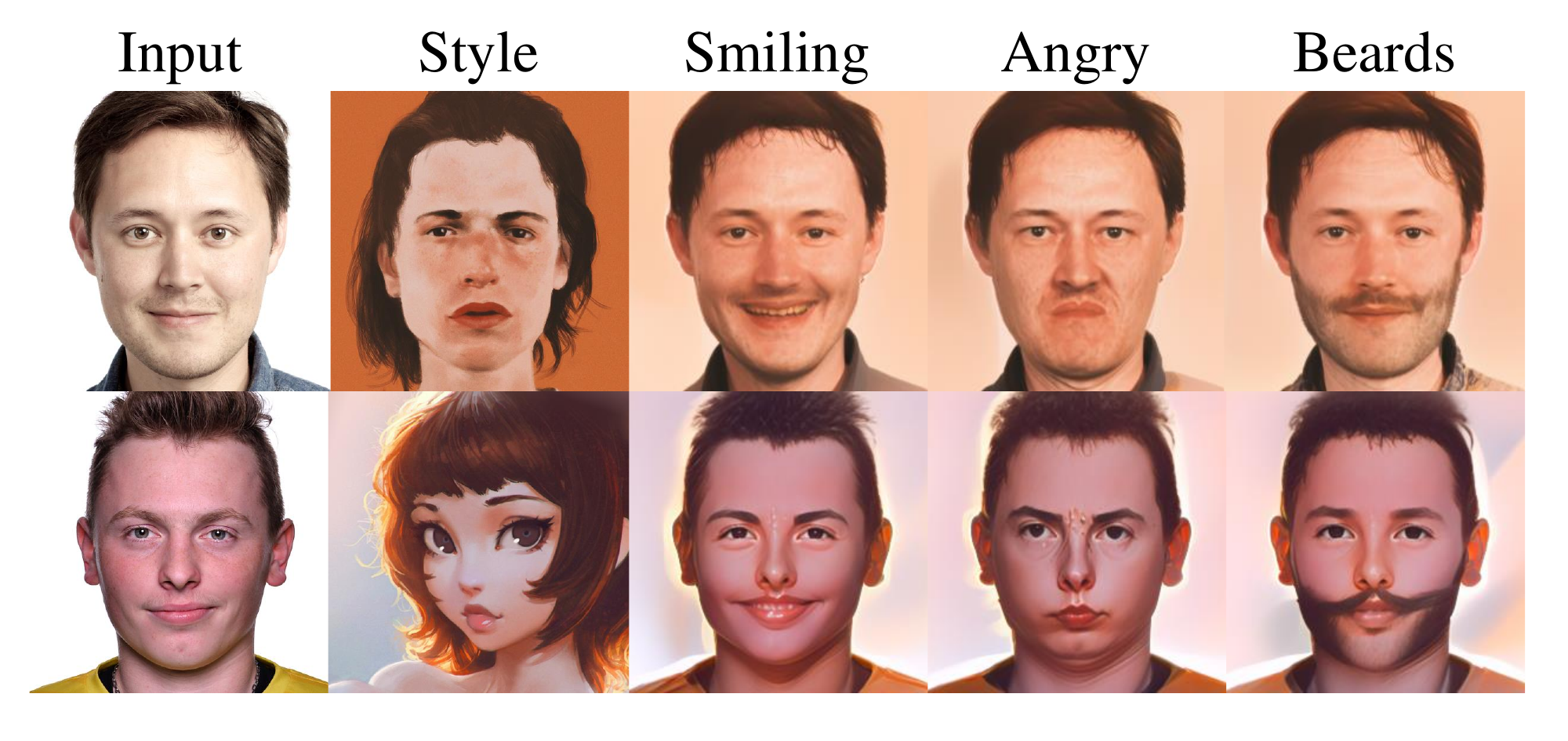}
    \vspace{-20pt}
    \caption{Stylization with text-driven manipulation. The optimized semantic code doesn't overwrite or harm the structure and style of the image, thus preserving the overall structure while manipulating attributes.}
    \vspace{-10pt}
    \label{fig:text_mani}
\end{figure}

\subsection{Stylization with Text-driven Manipulation}
\vspace{-5pt}
For text-driven manipulation, we directly optimize \zsemin using CLIP directional loss. Once \zsemin\ is optimized, it can be used to stylize the input image with the aforementioned mixing process. In Figure~\ref{fig:text_mani}, we show our qualitative results of stylization with text-driven manipulation, where our model successfully incorporates the style of a reference image with manipulated attributes while being robust in content preservation.

\begin{figure}[h]\centering
    \includegraphics[width=\linewidth]{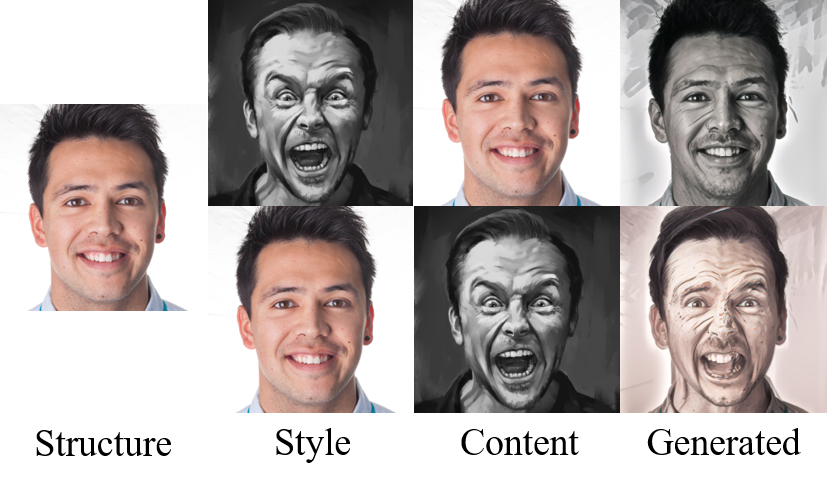}
    \vspace{-25pt}
    \caption{Ablation study of latent codes. The result shows that OSASIS is capable of effectively disentangling the structure from the semantics. By conditioning the semantic codes appropriately, we are able to control the content and style factors in the generated image.}
    \vspace{-10pt}
    \label{fig:ablation1}
\end{figure}

\begin{table}
    \centering
    \begin{tabular}{l|ccc}
    \hline
    & ArtFID$\downarrow $ & ID Sim$\uparrow$ & SD$\downarrow$ \\
    \hline
    w/o SPN & 36.41 & 0.6595 & 0.0371 \\
    \hline
    w SPN ($\lambda_{SPN}$=0.1) & \textbf{34.89} & 0.6825 & 0.0361 \\
    w SPN ($\lambda_{SPN}$=0.5) & 36.62 & 0.7177 & \textbf{0.0348} \\
    w SPN ($\lambda_{SPN}$=1.0) & 43.15 & \textbf{0.7321} & 0.0355 \\
    \hline
    \end{tabular}
    \vspace{-10pt}
    \caption{Quantitative ablation study of SPN}
    \vspace{-15pt}
    \label{table:abl_quant_comp}
\end{table}

\subsection{Ablation Study}
\vspace{-5pt}
\paragraph{Latent Code}
Furthermore, we conduct ablation studies to shed light on the nature of mixing content and style into the feature maps of the UNet. In the first row of Figure~\ref{fig:ablation1}, we perform normal stylization on an input image, \ie by encoding its structural latent code \xtin, conditioning \zsemstyle\ to low-level feature maps, and conditioning \zsemin\ to high-level feature maps. The second row shows the results of the semantic codes being conditioned oppositely. The resulting generated image 1) holds the structural integrity encoded from the structural latent code \xtin, 2) preserves the content from the given content image (\ie identity, facial expressions), and 3) retains the stylized attributes of the given style image (\ie skin complexion). We show that by conditioning the semantic codes to their respective feature maps, OSASIS achieves control over mixing content and style.

\vspace{-15pt}

\begin{figure}
    \centering
    \includegraphics[width=1.1\linewidth]{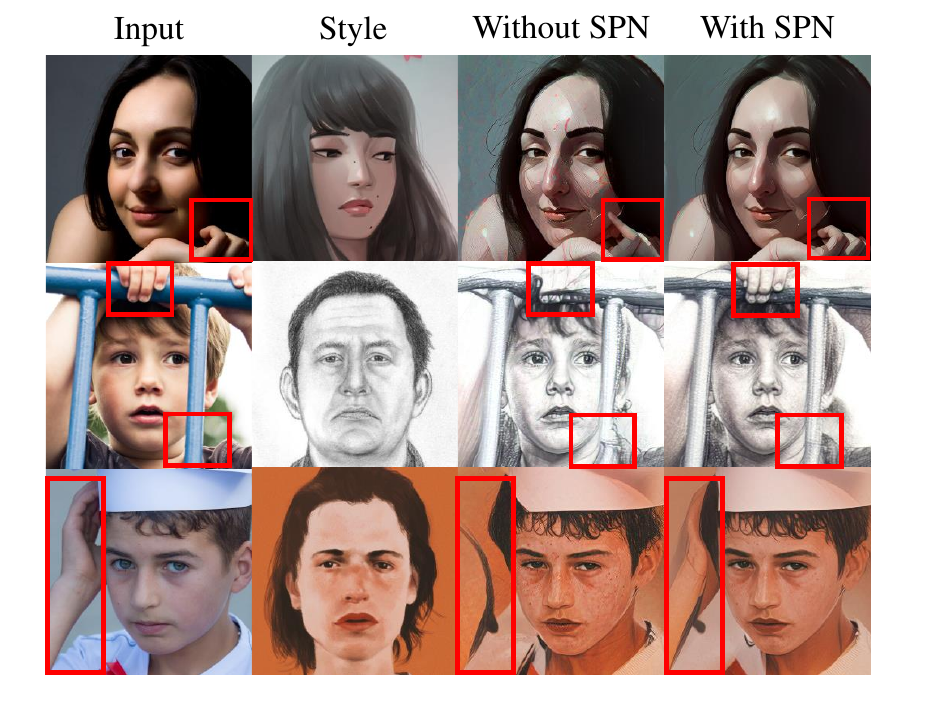}
    \vspace{-30pt}
    \caption{Ablation study of SPN. The results demonstrate that SPN is a crucial to ensure the preservation of the underlying structure while applying the stylization process.}
    \vspace{-15pt}
    \label{fig:ablation2}
\end{figure}

\paragraph{Structure-Preserving Network}
Our SPN is implemented to aid the structural latent code \xtin\ in preserving the overall structure of the stylized image, due to the encoding process Eq.\ref{eq:forward_diffae_in} being formulated to add noise. Figure~\ref{fig:ablation2} shows the effects of our SPN, where the third column is a stylized sample without, and the fourth column is a stylized sample with SPN. It can be seen that the stochastic latent code \xtin\ faithfully preserves the overall structure of the image, but without SPN, objects along the edges of content images (e.g. hands, poles, fingers) are distorted. To investigate the effect of SPN further, we conduct a quantitative ablation study. Table~\ref{table:abl_quant_comp} validates the efficacy of SPN, indicating substantial improvements in ID similarity and structure distance, thereby cementing its importance for maintaining content and structural integrity. It’s important to note, however, that a $\lambda_{SPN}$ value above 0.1 might overemphasize structural aspects, which can compromise the stylization quality, suggesting the necessity for careful calibration of $\lambda_{SPN}$ to achieve an optimal balance in stylization.

\vspace{-5pt}
\section{Related Work}
\vspace{-5pt}

\subsection{One-shot Image Stylization}
\vspace{-5pt}
Stylizing input images with only one reference image originates from neural style transfer, first introduced by Gatys \etal~\cite{neuralstyle}. However, this method requires the stylized image to be optimized every generation, which was addressed by Johnson \etal~\cite{johnson2016perceptual} by introducing an image transform network for fast stylization. However, traditional NST methods are limited in their ability to capture the semantic information of input and style images.

For GAN-based models, one-shot image stylization is achieved by using one-shot adaptation methods, which aims to transfer a generator to a new domain using a single reference image. One-shot adaptation methods typically involve finetuning a generator using only a single reference image. Once the generator is finetuned, these methods can unconditionally generate stylized images, and by using GAN inversion techniques, they also achieve input image stylization. StyleGAN-NADA~\cite{gal2022stylegan}, while applicable to one-shot image stylization tasks, exhibits limited capability, primarily due to its initial development for text-driven style transfer purposes. The first successful GAN-based one-shot adaptation method is MTG~\cite{zhu2021mind}, which uses CLIP directional loss to finetune the generator and mitigate overfitting problems. Other works have since focused on improving generation quality~\cite{kwon2022one}, content preservation~\cite{zhang2022towards}, and entity transfer~\cite{zhang2022generalized}. In contrast to one-shot adaptation approaches, our method aims to stylize real images with a single reference image instead of generating stylized synthesized images. Therefore, we refer to our method as one-shot image stylization. From our perspective, the work that is most comparable to our own is JoJoGAN~\cite{chong2022jojogan}. JoJoGAN generates a training dataset by random style mixing and finetunes a generator to create a style mapper.

\subsection{Image Manipulation with Diffusion Models}
\vspace{-5pt}
Image manipulation has advanced significantly in recent years, with methods presented by StyleGAN2~\cite{karras2020analyzing} being widely explored. However, the potential of diffusion models for high-quality image manipulation has been elucidated in recent research. DiffAE~\cite{preechakul2022diffusion} introduces a semantic encoder that generates semantically meaningful latent vectors for diffusion models, which can be manipulated for attribute editing. DiffusionCLIP~\cite{kim2022diffusionclip} demonstrates the effectiveness of diffusion-based text-guided image manipulation by fine-tuning a DDIM with CLIP directional loss. Asyrp~\cite{Asyrp} uncovers a semantic latent space in the architecture of diffusion models.The authors train the \textit{h-space} manipulation module with CLIP directional loss, achieving consistent image editing results. While DiffusionCLIP and Asyrp employ CLIP directional loss, their focus is on text-guided manipulation whereas our work targets image-guided manipulation. DiffuseIT~\cite{DiffuseIT} aims to perform image translation guided by either text or image using a CLIP and a pretrained ViT ~\cite{dosovitskiy2020image}. Their approach leverages the reverse process of DDPM and incorporates CLIP and ViT to guide the image generation process. InST~\cite{zhang2023inversion} employs textual inversion to extract the concept from a style image. By conditioning the generation process on this extracted concept, InST is able to stylize input images.
\vspace{-5pt}
\section{Conclusion}
\vspace{-5pt}
We have introduced OSASIS, a novel one-shot image stylization method based on diffusion models. In contrast to GAN-based and other diffusion-based stylization methods, OSASIS shows robust structure awareness in stylization, effectively disentangling the structure and semantics from an image. While OSASIS demonstrates significant advancements in structure-aware stylization, several limitations exist. A notable constraint of OSASIS is its training time, which is longer than comparison methods. This extended training duration is a trade-off for the method’s enhanced ability to maintain structural integrity and adapt to various styles. Additionally, OSASIS requires training for each style image. This requirement can be seen as a limitation in scenarios where rapid deployment across multiple styles is needed. Despite these challenges, the robustness of OSASIS in preserving the structural integrity of the input images, its effectiveness in out-of-domain reference stylization, and its adaptability in text-driven manipulation make it a promising approach in the field of stylized image synthesis. Future work will address these limitations, particularly in optimizing training efficiency and reducing the necessity for individual style image training, to enhance the practicality and applicability of OSASIS in diverse real-world scenarios.

\vspace{-10pt}
\paragraph{Acknowledgment}
We thank the ImageVision team of NAVER Cloud for their thoughtful advice and discussions. Training and experiments were done on the Naver Smart Machine Learning (NSML) platform~\cite{kim2018nsml}. This study was supported by BK21 FOUR.

{
    \small
    \bibliographystyle{ieeenat_fullname}
    \bibliography{main}

\begin{thebibliography}{45}
\providecommand{\natexlab}[1]{#1}
\providecommand{\url}[1]{\texttt{#1}}
\expandafter\ifx\csname urlstyle\endcsname\relax
  \providecommand{\doi}[1]{doi: #1}\else
  \providecommand{\doi}{doi: \begingroup \urlstyle{rm}\Url}\fi

\bibitem[Alaluf et~al.(2021)Alaluf, Patashnik, and Cohen-Or]{alaluf2021restyle}
Yuval Alaluf, Or Patashnik, and Daniel Cohen-Or.
\newblock Restyle: A residual-based stylegan encoder via iterative refinement.
\newblock In \emph{Proceedings of the IEEE/CVF International Conference on Computer Vision}, pages 6711--6720, 2021.

\bibitem[Choi et~al.(2020)Choi, Uh, Yoo, and Ha]{choi2020stargan}
Yunjey Choi, Youngjung Uh, Jaejun Yoo, and Jung-Woo Ha.
\newblock Stargan v2: Diverse image synthesis for multiple domains.
\newblock In \emph{Proceedings of the IEEE/CVF conference on computer vision and pattern recognition}, pages 8188--8197, 2020.

\bibitem[Chong and Forsyth(2022)]{chong2022jojogan}
Min~Jin Chong and David Forsyth.
\newblock Jojogan: One shot face stylization.
\newblock In \emph{European Conference on Computer Vision}, pages 128--152. Springer, 2022.

\bibitem[Deng et~al.(2019)Deng, Guo, Xue, and Zafeiriou]{deng2019arcface}
Jiankang Deng, Jia Guo, Niannan Xue, and Stefanos Zafeiriou.
\newblock Arcface: Additive angular margin loss for deep face recognition.
\newblock In \emph{Proceedings of the IEEE/CVF conference on computer vision and pattern recognition}, pages 4690--4699, 2019.

\bibitem[Dhariwal and Nichol(2021)]{dhariwal2021diffusion}
Prafulla Dhariwal and Alexander Nichol.
\newblock Diffusion models beat gans on image synthesis.
\newblock \emph{Advances in Neural Information Processing Systems}, 34:\penalty0 8780--8794, 2021.

\bibitem[Dosovitskiy et~al.(2020)Dosovitskiy, Beyer, Kolesnikov, Weissenborn, Zhai, Unterthiner, Dehghani, Minderer, Heigold, Gelly, et~al.]{dosovitskiy2020image}
Alexey Dosovitskiy, Lucas Beyer, Alexander Kolesnikov, Dirk Weissenborn, Xiaohua Zhai, Thomas Unterthiner, Mostafa Dehghani, Matthias Minderer, Georg Heigold, Sylvain Gelly, et~al.
\newblock An image is worth 16x16 words: Transformers for image recognition at scale.
\newblock \emph{arXiv preprint arXiv:2010.11929}, 2020.

\bibitem[Gal et~al.(2022)Gal, Patashnik, Maron, Bermano, Chechik, and Cohen-Or]{gal2022stylegan}
Rinon Gal, Or Patashnik, Haggai Maron, Amit~H Bermano, Gal Chechik, and Daniel Cohen-Or.
\newblock Stylegan-nada: Clip-guided domain adaptation of image generators.
\newblock \emph{ACM Transactions on Graphics (TOG)}, 41\penalty0 (4):\penalty0 1--13, 2022.

\bibitem[Gatys et~al.(2016)Gatys, Ecker, and Bethge]{neuralstyle}
Leon~A Gatys, Alexander~S Ecker, and Matthias Bethge.
\newblock Image style transfer using convolutional neural networks.
\newblock In \emph{Proceedings of the IEEE conference on computer vision and pattern recognition}, pages 2414--2423, 2016.

\bibitem[Ho and Salimans(2021)]{ho2021classifier}
Jonathan Ho and Tim Salimans.
\newblock Classifier-free diffusion guidance.
\newblock In \emph{NeurIPS 2021 Workshop on Deep Generative Models and Downstream Applications}, 2021.

\bibitem[Ho et~al.(2020)Ho, Jain, and Abbeel]{ho2020denoising}
Jonathan Ho, Ajay Jain, and Pieter Abbeel.
\newblock Denoising diffusion probabilistic models.
\newblock \emph{Advances in Neural Information Processing Systems}, 33:\penalty0 6840--6851, 2020.

\bibitem[Johnson et~al.(2016)Johnson, Alahi, and Fei-Fei]{johnson2016perceptual}
Justin Johnson, Alexandre Alahi, and Li Fei-Fei.
\newblock Perceptual losses for real-time style transfer and super-resolution.
\newblock In \emph{Computer Vision--ECCV 2016: 14th European Conference, Amsterdam, The Netherlands, October 11-14, 2016, Proceedings, Part II 14}, pages 694--711. Springer, 2016.

\bibitem[Karras et~al.(2019)Karras, Laine, and Aila]{karras2019style}
Tero Karras, Samuli Laine, and Timo Aila.
\newblock A style-based generator architecture for generative adversarial networks.
\newblock In \emph{Proceedings of the IEEE/CVF conference on computer vision and pattern recognition}, pages 4401--4410, 2019.

\bibitem[Karras et~al.(2020{\natexlab{a}})Karras, Aittala, Hellsten, Laine, Lehtinen, and Aila]{karras2020training}
Tero Karras, Miika Aittala, Janne Hellsten, Samuli Laine, Jaakko Lehtinen, and Timo Aila.
\newblock Training generative adversarial networks with limited data.
\newblock \emph{Advances in neural information processing systems}, 33:\penalty0 12104--12114, 2020{\natexlab{a}}.

\bibitem[Karras et~al.(2020{\natexlab{b}})Karras, Laine, Aittala, Hellsten, Lehtinen, and Aila]{karras2020analyzing}
Tero Karras, Samuli Laine, Miika Aittala, Janne Hellsten, Jaakko Lehtinen, and Timo Aila.
\newblock Analyzing and improving the image quality of stylegan.
\newblock In \emph{Proceedings of the IEEE/CVF conference on computer vision and pattern recognition}, pages 8110--8119, 2020{\natexlab{b}}.

\bibitem[Kim et~al.(2022)Kim, Kwon, and Ye]{kim2022diffusionclip}
Gwanghyun Kim, Taesung Kwon, and Jong~Chul Ye.
\newblock Diffusionclip: Text-guided diffusion models for robust image manipulation.
\newblock In \emph{Proceedings of the IEEE/CVF Conference on Computer Vision and Pattern Recognition}, pages 2426--2435, 2022.

\bibitem[Kim et~al.(2018)Kim, Kim, Seo, Kim, Park, Park, Jo, Kim, Yang, Kim, et~al.]{kim2018nsml}
Hanjoo Kim, Minkyu Kim, Dongjoo Seo, Jinwoong Kim, Heungseok Park, Soeun Park, Hyunwoo Jo, KyungHyun Kim, Youngil Yang, Youngkwan Kim, et~al.
\newblock Nsml: Meet the mlaas platform with a real-world case study.
\newblock \emph{arXiv preprint arXiv:1810.09957}, 2018.

\bibitem[Kwon and Ye(2022{\natexlab{a}})]{DiffuseIT}
Gihyun Kwon and Jong~Chul Ye.
\newblock Diffusion-based image translation using disentangled style and content representation.
\newblock \emph{arXiv preprint arXiv:2209.15264}, 2022{\natexlab{a}}.

\bibitem[Kwon and Ye(2022{\natexlab{b}})]{kwon2022one}
Gihyun Kwon and Jong~Chul Ye.
\newblock One-shot adaptation of gan in just one clip.
\newblock \emph{arXiv preprint arXiv:2203.09301}, 2022{\natexlab{b}}.

\bibitem[Kwon et~al.(2022)Kwon, Jeong, and Uh]{Asyrp}
Mingi Kwon, Jaeseok Jeong, and Youngjung Uh.
\newblock Diffusion models already have a semantic latent space.
\newblock \emph{arXiv preprint arXiv:2210.10960}, 2022.

\bibitem[Liu et~al.(2021)Liu, Li, Qin, Zhang, Wan, and Zheng]{liu2021blendgan}
Mingcong Liu, Qiang Li, Zekui Qin, Guoxin Zhang, Pengfei Wan, and Wen Zheng.
\newblock Blendgan: Implicitly gan blending for arbitrary stylized face generation.
\newblock \emph{Advances in Neural Information Processing Systems}, 34:\penalty0 29710--29722, 2021.

\bibitem[Liu et~al.(2016)Liu, Luo, Qiu, Wang, and Tang]{liuLQWTcvpr16DeepFashion}
Ziwei Liu, Ping Luo, Shi Qiu, Xiaogang Wang, and Xiaoou Tang.
\newblock Deepfashion: Powering robust clothes recognition and retrieval with rich annotations.
\newblock In \emph{Proceedings of IEEE Conference on Computer Vision and Pattern Recognition (CVPR)}, 2016.

\bibitem[Meng et~al.(2022)Meng, He, Song, Song, Wu, Zhu, and Ermon]{meng2022sdedit}
Chenlin Meng, Yutong He, Yang Song, Jiaming Song, Jiajun Wu, Jun-Yan Zhu, and Stefano Ermon.
\newblock {SDE}dit: Guided image synthesis and editing with stochastic differential equations.
\newblock In \emph{International Conference on Learning Representations}, 2022.

\bibitem[Ojha et~al.(2021)Ojha, Li, Lu, Efros, Lee, Shechtman, and Zhang]{ojha2021few}
Utkarsh Ojha, Yijun Li, Jingwan Lu, Alexei~A Efros, Yong~Jae Lee, Eli Shechtman, and Richard Zhang.
\newblock Few-shot image generation via cross-domain correspondence.
\newblock In \emph{Proceedings of the IEEE/CVF Conference on Computer Vision and Pattern Recognition}, pages 10743--10752, 2021.

\bibitem[Pinkney and Adler(2020)]{pinkney2020resolution}
Justin~NM Pinkney and Doron Adler.
\newblock Resolution dependent gan interpolation for controllable image synthesis between domains.
\newblock \emph{arXiv preprint arXiv:2010.05334}, 2020.

\bibitem[Preechakul et~al.(2022)Preechakul, Chatthee, Wizadwongsa, and Suwajanakorn]{preechakul2022diffusion}
Konpat Preechakul, Nattanat Chatthee, Suttisak Wizadwongsa, and Supasorn Suwajanakorn.
\newblock Diffusion autoencoders: Toward a meaningful and decodable representation.
\newblock In \emph{Proceedings of the IEEE/CVF Conference on Computer Vision and Pattern Recognition}, pages 10619--10629, 2022.

\bibitem[Ramachandran et~al.(2017)Ramachandran, Zoph, and Le]{ramachandran2017searching}
Prajit Ramachandran, Barret Zoph, and Quoc~V Le.
\newblock Searching for activation functions.
\newblock \emph{arXiv preprint arXiv:1710.05941}, 2017.

\bibitem[Roich et~al.(2022)Roich, Mokady, Bermano, and Cohen-Or]{roich2022pivotal}
Daniel Roich, Ron Mokady, Amit~H Bermano, and Daniel Cohen-Or.
\newblock Pivotal tuning for latent-based editing of real images.
\newblock \emph{ACM Transactions on graphics (TOG)}, 42\penalty0 (1):\penalty0 1--13, 2022.

\bibitem[Rombach et~al.(2022)Rombach, Blattmann, Lorenz, Esser, and Ommer]{rombach2022high}
Robin Rombach, Andreas Blattmann, Dominik Lorenz, Patrick Esser, and Bj{\"o}rn Ommer.
\newblock High-resolution image synthesis with latent diffusion models.
\newblock In \emph{Proceedings of the IEEE/CVF Conference on Computer Vision and Pattern Recognition}, pages 10684--10695, 2022.

\bibitem[Saharia et~al.(2022{\natexlab{a}})Saharia, Chan, Saxena, Li, Whang, Denton, Ghasemipour, Ayan, Mahdavi, Lopes, et~al.]{saharia2022photorealistic}
Chitwan Saharia, William Chan, Saurabh Saxena, Lala Li, Jay Whang, Emily Denton, Seyed Kamyar~Seyed Ghasemipour, Burcu~Karagol Ayan, S~Sara Mahdavi, Rapha~Gontijo Lopes, et~al.
\newblock Photorealistic text-to-image diffusion models with deep language understanding.
\newblock \emph{arXiv preprint arXiv:2205.11487}, 2022{\natexlab{a}}.

\bibitem[Saharia et~al.(2022{\natexlab{b}})Saharia, Ho, Chan, Salimans, Fleet, and Norouzi]{saharia2022image}
Chitwan Saharia, Jonathan Ho, William Chan, Tim Salimans, David~J Fleet, and Mohammad Norouzi.
\newblock Image super-resolution via iterative refinement.
\newblock \emph{IEEE Transactions on Pattern Analysis and Machine Intelligence}, 2022{\natexlab{b}}.

\bibitem[Song et~al.(2021)Song, Luo, Liu, Ma, Lai, Zheng, and Cham]{song2021agilegan}
Guoxian Song, Linjie Luo, Jing Liu, Wan-Chun Ma, Chunpong Lai, Chuanxia Zheng, and Tat-Jen Cham.
\newblock Agilegan: stylizing portraits by inversion-consistent transfer learning.
\newblock \emph{ACM Transactions on Graphics (TOG)}, 40\penalty0 (4):\penalty0 1--13, 2021.

\bibitem[Song et~al.(2020)Song, Meng, and Ermon]{song2020denoising}
Jiaming Song, Chenlin Meng, and Stefano Ermon.
\newblock Denoising diffusion implicit models.
\newblock In \emph{International Conference on Learning Representations}, 2020.

\bibitem[Tov et~al.(2021)Tov, Alaluf, Nitzan, Patashnik, and Cohen-Or]{tov2021designing}
Omer Tov, Yuval Alaluf, Yotam Nitzan, Or Patashnik, and Daniel Cohen-Or.
\newblock Designing an encoder for stylegan image manipulation.
\newblock \emph{ACM Transactions on Graphics (TOG)}, 40\penalty0 (4):\penalty0 1--14, 2021.

\bibitem[Tumanyan et~al.(2022)Tumanyan, Bar-Tal, Bagon, and Dekel]{tumanyan2022splicing}
Narek Tumanyan, Omer Bar-Tal, Shai Bagon, and Tali Dekel.
\newblock Splicing vit features for semantic appearance transfer.
\newblock In \emph{Proceedings of the IEEE/CVF Conference on Computer Vision and Pattern Recognition}, pages 10748--10757, 2022.

\bibitem[Vaswani et~al.(2017)Vaswani, Shazeer, Parmar, Uszkoreit, Jones, Gomez, Kaiser, and Polosukhin]{vaswani2017attention}
Ashish Vaswani, Noam Shazeer, Niki Parmar, Jakob Uszkoreit, Llion Jones, Aidan~N Gomez, {\L}ukasz Kaiser, and Illia Polosukhin.
\newblock Attention is all you need.
\newblock \emph{Advances in neural information processing systems}, 30, 2017.

\bibitem[Wang et~al.(2022)Wang, Zhang, Fan, Wang, and Chen]{wang2022high}
Tengfei Wang, Yong Zhang, Yanbo Fan, Jue Wang, and Qifeng Chen.
\newblock High-fidelity gan inversion for image attribute editing.
\newblock In \emph{Proceedings of the IEEE/CVF Conference on Computer Vision and Pattern Recognition}, 2022.

\bibitem[Wright and Ommer(2022)]{wright2022artfid}
Matthias Wright and Bj{\"o}rn Ommer.
\newblock Artfid: Quantitative evaluation of neural style transfer.
\newblock In \emph{Pattern Recognition: 44th DAGM German Conference, DAGM GCPR 2022, Konstanz, Germany, September 27--30, 2022, Proceedings}, pages 560--576. Springer, 2022.

\bibitem[Wu and He(2018)]{wu2018group}
Yuxin Wu and Kaiming He.
\newblock Group normalization.
\newblock In \emph{Proceedings of the European conference on computer vision (ECCV)}, pages 3--19, 2018.

\bibitem[Xu et~al.(2023)Xu, Shu, Smith, Oh, and Huang]{xu2023n}
Yiran Xu, Zhixin Shu, Cameron Smith, Seoung~Wug Oh, and Jia-Bin Huang.
\newblock In-n-out: Faithful 3d gan inversion with volumetric decomposition for face editing.
\newblock \emph{arXiv preprint arXiv:2302.04871}, 2023.

\bibitem[Yu et~al.(2015)Yu, Seff, Zhang, Song, Funkhouser, and Xiao]{yu2015lsun}
Fisher Yu, Ari Seff, Yinda Zhang, Shuran Song, Thomas Funkhouser, and Jianxiong Xiao.
\newblock Lsun: Construction of a large-scale image dataset using deep learning with humans in the loop.
\newblock \emph{arXiv preprint arXiv:1506.03365}, 2015.

\bibitem[Zhang et~al.(2018)Zhang, Isola, Efros, Shechtman, and Wang]{zhang2018unreasonable}
Richard Zhang, Phillip Isola, Alexei~A Efros, Eli Shechtman, and Oliver Wang.
\newblock The unreasonable effectiveness of deep features as a perceptual metric.
\newblock In \emph{Proceedings of the IEEE conference on computer vision and pattern recognition}, pages 586--595, 2018.

\bibitem[Zhang et~al.(2022{\natexlab{a}})Zhang, Wei, Ji, Bai, Zuo, et~al.]{zhang2022towards}
Yabo Zhang, Yuxiang Wei, Zhilong Ji, Jinfeng Bai, Wangmeng Zuo, et~al.
\newblock Towards diverse and faithful one-shot adaption of generative adversarial networks.
\newblock In \emph{Advances in Neural Information Processing Systems}, 2022{\natexlab{a}}.

\bibitem[Zhang et~al.(2023)Zhang, Huang, Tang, Huang, Ma, Dong, and Xu]{zhang2023inversion}
Yuxin Zhang, Nisha Huang, Fan Tang, Haibin Huang, Chongyang Ma, Weiming Dong, and Changsheng Xu.
\newblock Inversion-based style transfer with diffusion models.
\newblock In \emph{Proceedings of the IEEE/CVF Conference on Computer Vision and Pattern Recognition}, pages 10146--10156, 2023.

\bibitem[Zhang et~al.(2022{\natexlab{b}})Zhang, Liu, Han, Guo, Yao, and Mei]{zhang2022generalized}
Zicheng Zhang, Yinglu Liu, Congying Han, Tiande Guo, Ting Yao, and Tao Mei.
\newblock Generalized one-shot domain adaption of generative adversarial networks.
\newblock \emph{arXiv preprint arXiv:2209.03665}, 2022{\natexlab{b}}.

\bibitem[Zhu et~al.(2021)Zhu, Abdal, Femiani, and Wonka]{zhu2021mind}
Peihao Zhu, Rameen Abdal, John Femiani, and Peter Wonka.
\newblock Mind the gap: Domain gap control for single shot domain adaptation for generative adversarial networks.
\newblock In \emph{International Conference on Learning Representations}, 2021.

\end{thebibliography}
}

\newpage
\clearpage

\setcounter{table}{2}
\setcounter{figure}{9}
\setcounter{equation}{14}

\newpage
\appendix

\twocolumn[
\centering
\Large
\textbf{One-Shot Structure-Aware Stylized Image Synthesis} \\
\vspace{0.5em}-- Supplementary Material -- \\
\vspace{15pt}
]

\section{Training Details}
\subsection{Training Preparation}
In order to train one-shot structure-aware stylized image synthesis (OSASIS), it is necessary to create a photorealistic image, denoted as \IstyleA, that is semantically aligned with a given style image, \IstyleB. This is achieved by first encoding \IstyleB\ into a latent code, represented as \xtzero\ using Eq.9. Then \IstyleA\ is generated from \xtzero\ using Eq.10, utilizing a pretrained DDPM $\epsilon_\theta$ during its encoding and generation phases. However, since these processes are stochastic in nature, the resulting \IstyleA\ may not always align perfectly with \IstyleB. To overcome this, we generate 30 images from \IstyleB\ and evaluate their alignment with \IstyleB using both the $L_1$ loss and perceptual similarity loss~\cite{zhang2018unreasonable}. We then select the image that is most similar to \IstyleB\ as \IstyleA\ as the prime candidate. In cases of out-of-domain (OOD) reference images, we match the domain of the pretrained DDPM with the domain of the style image. (\eg for church style images, we use a pretrained DDPM trained on LSUN-church).

\subsection{Structure-Preserving Network}
\begin{figure}\centering
    \includegraphics[width=0.9\linewidth]{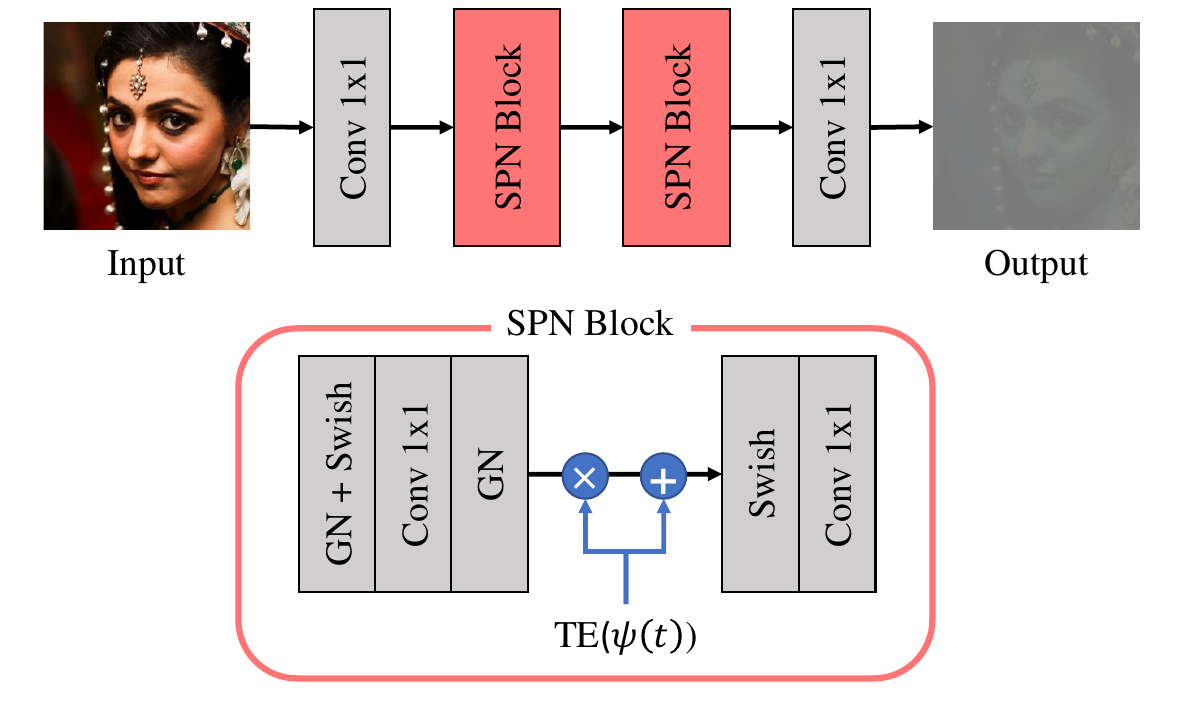}
    \caption{The SPN architecture is comprised of SPN blocks, each consisting of two 1x1 convolutions with group normalizations~\cite{wu2018group} and swish activations~\cite{ramachandran2017searching}. To incorporate the temporal information of each timestep, the SPN block employs sinusoidal position embeddings~\cite{vaswani2017attention}, represented as $\psi$ for each time step. Timestep embedding (TE) layers are also incorporated, consisting of two linear layers. To condition the SPN block on the current timestep, we use the timestep embedding as a scale and shift parameter of group normalization, which is similar to previous works~\cite{dhariwal2021diffusion}.}
    \vspace{-10pt}
    \label{fig:SPN}
\end{figure}

The structure-preserving network (SPN) incorporates a 1x1 convolution to preserve the overall structure of the input image. During the reverse process, the SPN's output is integrated with the DDIM output. Given that this integration takes place at every timestep, it is crucial for the network to recognize the timestep to effectively control structure preservation. To facilitate this, each block of the SPN is conditioned on the timestep. The detailed architecture of the SPN is illustrated in Figure~\ref{fig:SPN}.

\subsection{Loss Function Formulation} \label{supp:loss}
\paragraph{Cross-Domain Loss}
The objective of the cross-domain loss is to align the directional shifts from domain A to domain B, ensuring that the change from \IinA\ to \IinB\ is kept consistent with the change from \IstyleA\ to \IstyleB. Leveraging the CLIP image encoder $E_I$, latent vectors of both the input and style images are extracted. Subtracting these semantically aligned latent vectors results in semantically meaningful directions. The changes in the style and input images are calculated using Eq.\ref{eq:v_style} and Eq.\ref{eq:v_in}, respectively. The cosine similarity, as described in Eq.~\ref{eq:cross-dom loss}, is then used to evaluate the similarity of the two directions.
\vspace{-5pt}
\begin{gather}
    \label{eq:v_style}
    \mathbf{v}_{\mathrm{style}} = E_I(I^{\mathrm{style}}_{B}) - E_I(I^{\mathrm{style}}_{A})\\
    \label{eq:v_in}
    \mathbf{v}_{\mathrm{in}} = E_I(I^{\mathrm{in}}_{B}) - E_I(I^{\mathrm{in}}_{A})\\
    \label{eq:cross-dom loss}
    L_{cross} = 1 - sim(\mathbf{v}_{style}, \mathbf{v}_{in})
\end{gather}

\paragraph{In-Domain Loss}
The purpose of the in-domain loss is to mitigate unintended changes in the direction of stylization, which can often result in excessive reflection of the style image. This is achieved by measuring the similarity of changes within both domains A and B. Like the cross-domain loss, the in-domain changes are calculated using Eq.\ref{eq:v_A} and Eq.\ref{eq:v_B}. The similarity between the two directions is then determined using Eq.~\ref{eq:in-dom loss}.
\vspace{-5pt}
\begin{gather}
    \label{eq:v_A}
    \mathbf{v}_{A} = E_I(I^{\mathrm{cont}}_{A}) - E_I(I^{\mathrm{style}}_{A})\\
    \label{eq:v_B}
    \mathbf{v}_{B} = E_I(I^{\mathrm{cont}}_{B}) - E_I(I^{\mathrm{style}}_{B})\\
    \label{eq:in-dom loss}
    L_{In} = 1 - sim(\mathbf{v}_{A}, \mathbf{v}_{B})
\end{gather}

\paragraph{Reconstruction Loss}
The reconstruction loss aims to guarantee that the photorealistic style image \IstyleA\ can be accurately translated from domain A to domain B. This is achieved by encoding \IstyleA\ using \DiffA, which results in the latent vectors \zsemstyle\ and \xtstyle. The latent vectors are then fed into \DiffB, which generates the predicted domain B style image $\hat{I}^{\mathrm{style}}_B$. The reconstruction loss is calculated by comparing $\hat{I}^{\mathrm{style}}_B$ with \IstyleB\ and the summation of the $L_1$ loss, perceptual similarity loss~\cite{zhang2018unreasonable}, and the $L_1$ CLIP embedding loss described in Eq.~\ref{eq: recon image loss}-\ref{eq: recon clip loss}. The total reconstruct loss can be computed using Eq.~\ref{eq: recon loss}.
\vspace{-5pt}
\begin{gather}
    \label{eq: image hat}
    \hat{I^{\mathrm{style}}_{B}} = \epsilon_{\theta}^{B}(\mathbf{z}_{\mathrm{sem}}^{\mathrm{style}}, \mathbf{x}_{T}^{\mathrm{style}})\\
    \label{eq: recon image loss}
    L_{re-image} = L_1(I^{\mathrm{style}}_{B}, \hat{I^{\mathrm{style}}_{B}})\\
    \label{eq: recon lpips loss}
    L_{re-lpips} = L_{\mathrm{lpips}}(I^{\mathrm{style}}_{B}, \hat{I^{\mathrm{style}}_{B}})\\
    \label{eq: recon clip loss}
    L_{re-clip} = L_1(E_I(I^{\mathrm{style}}_{B}), E_I(\hat{I^{\mathrm{style}}_{B}}))\\
    L_{recon} = \lambda_{re-image}L_{re-image} + \lambda_{re-lpips}L_{re-lpips} \nonumber \\
    \label{eq: recon loss}
    + \lambda_{re-clip}L_{re-clip}
\end{gather}

\paragraph{Total Loss}
The total loss is a weighted sum of the aforementioned cross-domain, in-domain, and reconstruction loss, formulated by the following equation:
\begin{gather}
    \label{eq: total loss}
    L_{total} = \lambda_{cross}L_{cross} + \lambda_{in}L_{in} + L_{recon}
\end{gather}

\section{Sampling Details}

\subsection{Mixing Content and Style} \label{supp:mix_cont_style}
\begin{figure*}\centering
    \includegraphics[width=0.9\linewidth]{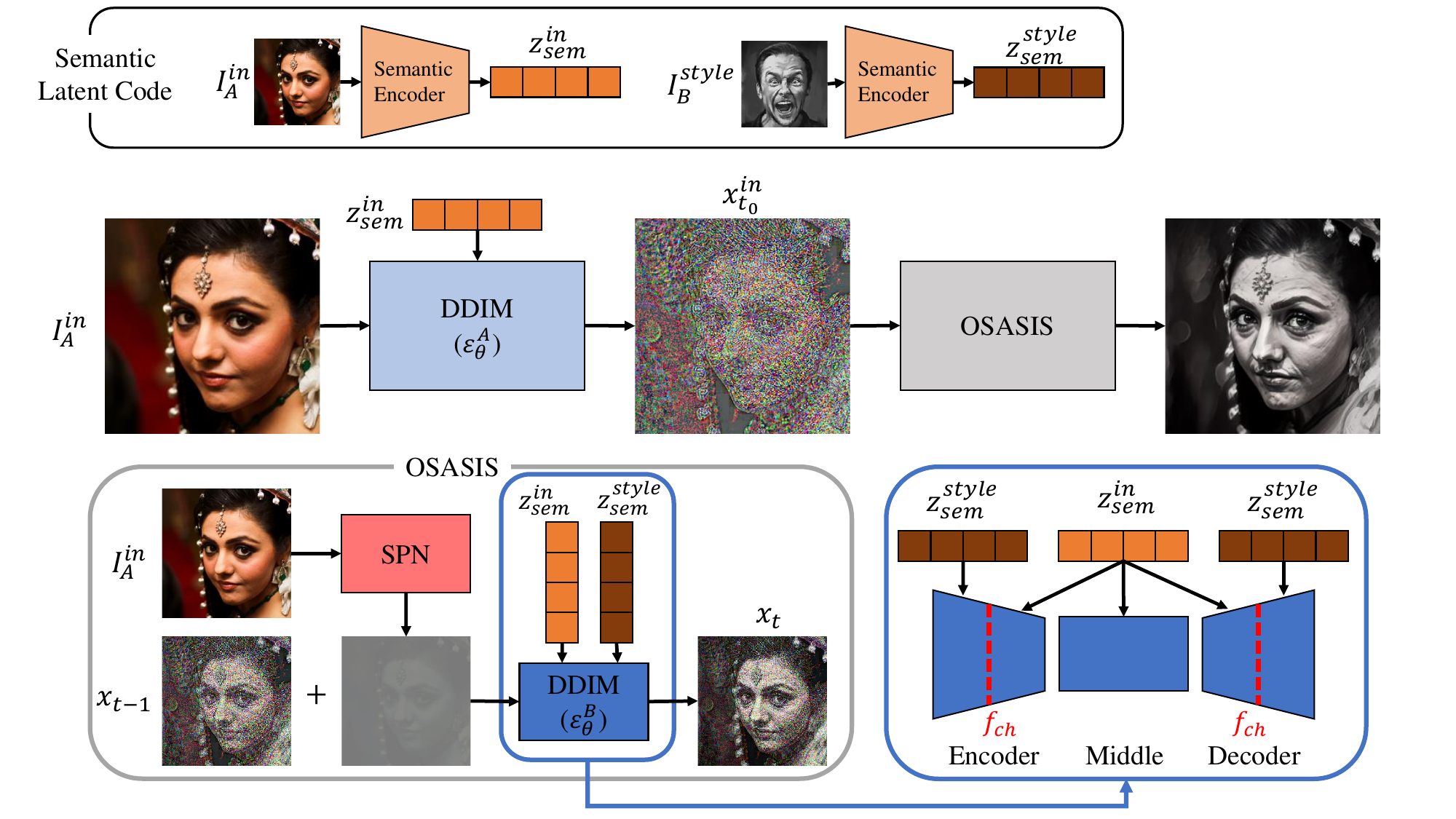}
    \vspace{-10pt}
    \caption{Sampling process of OSASIS. In contrast to the training phase, during the sampling process, \zsemstyle\ is derived from \IstyleB. To integrate the content from the input with the style of the style image, we condition the DDIM \DiffB, on \zsemstyle\ and \zsemin, respectively. This conditioning is separated by $f_{ch}$.}
    \vspace{-10pt}
    \label{fig:sampling}
\end{figure*}

After training the DDIM \DiffB, we can combine the content of the input images with the style of style images. We achieve this by encoding these images into semantic latent codes, specifically \zsemin\ for input and \zsemstyle for style. It is important to note that during training, \zsemstyle\ comes from \IstyleA, but for sampling, it is sourced from \IstyleB. Using the pretrained DDIM \DiffA, we encode \IinA\ into a structural latent code \xtin. From \xtin, \DiffB\ generates a stylized image. The process of generating a stylized image is similar to the process of generating \IinB\ from \IinA, as described in Eq.12-14. However, the sampling step involves two semantic latent codes, \zsemin\ and \zsemstyle, so the generation process is adjusted accordingly:
\vspace{-5pt}
\begin{gather}
    \mathbf{x}_{t-1} = \sqrt{\alpha_{t-1}}f_\theta(\mathbf{x}'_{t}, t, \mathbf{z}_{\mathrm{sem}}^{\mathrm{in}}, \mathbf{z}_{\mathrm{sem}}^{\mathrm{style}}) + \nonumber \\
    \label{eq:reverse_diffae_in_sample}
    \sqrt{1 - \alpha_{t-1}}\epsilon^B_{\theta}(\mathbf{x}_{t}', t, \mathbf{z}_{\mathrm{sem}}^{\mathrm{in}}, \mathbf{z}_{\mathrm{sem}}^{\mathrm{style}})
\end{gather}
\zsemstyle\ is conditioned on low-level feature maps, while \zsemin\ is conditioned on high-level feature maps. The separation between these feature maps is done using $f_{ch}$. The overall sampling process is illustrated in Figure~\ref{fig:sampling}.

\subsection{Text-driven Manipulation} \label{supp:text_mani}
To enable text-driven image manipulation, we utilize CLIP directional loss, as in previous works~\cite{kim2022diffusionclip}. The CLIP directional loss aligns the source-to-target text change with the source-to-target image change, similar to the cross-domain loss. The CLIP text encoder is denoted as $E_T$, and the target and source text is represented as $T_{\mathrm{trg}}$ and $T_{\mathrm{src}}$, respectively. The source image \IinA\ is encoded into semantic and structural latent codes, \zsemin\ and \xtin, using DDIM \DiffA. We freeze \xtin\ and \DiffA, and optimize \zsemin\ to obtain the optimal semantic latent code $\mathbf{z}_{\mathrm{sem}}^{\mathrm{in}*}$. 
Following this, employing \xtin\ and $\mathbf{z}_{\mathrm{sem}}^{\mathrm{in}*}$ allows us to derive $I^{\mathrm{in}}_{opt}$ utilizing DDIM \DiffB. The CLIP directional loss is computed as follows:
\vspace{-5pt}
\begin{gather}
    \label{eq:v_text}
    \mathbf{v}_{\mathrm{text}} = E_T(T_{\mathrm{trg}}) - E_T(T_{\mathrm{src}})\\
    \label{eq:v_image}
    \mathbf{v}_{\mathrm{image}} = E_I(I^{\mathrm{in}}_{opt}) - E_I(I^{\mathrm{in}}_{A})\\
    \label{eq:text loss}
    L_{text} = 1 - sim(\mathbf{v}_{text}, \mathbf{v}_{image})
\end{gather}

\begin{figure}\centering
    \includegraphics[width=\linewidth]{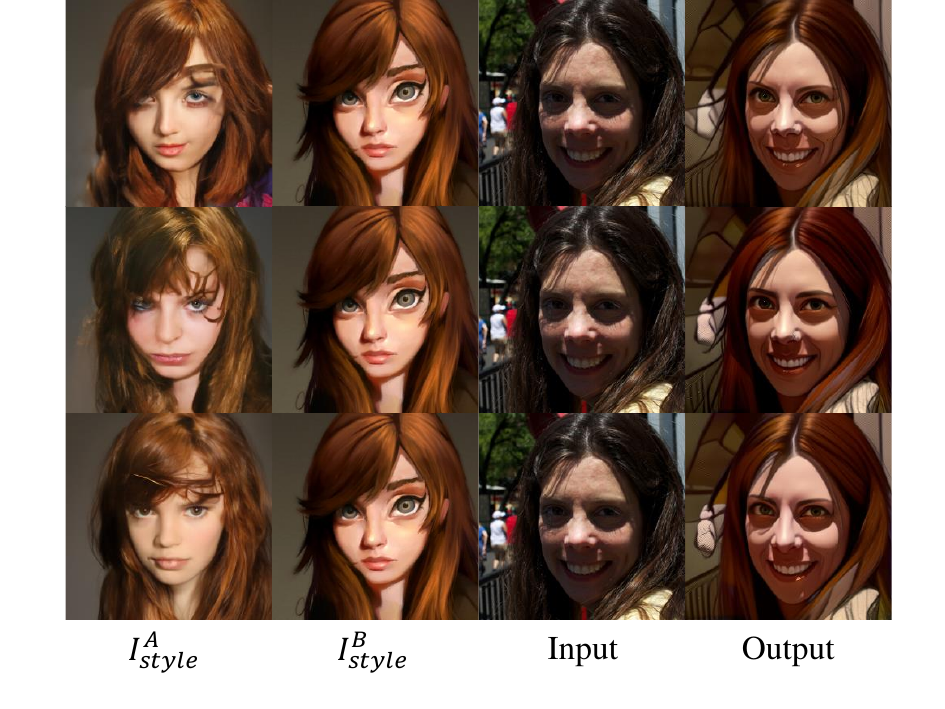}
    \vspace{-25pt}
    \caption{Reliance on \IstyleA. The representation of \IstyleA is stochastic, it can be adjusted through the sampling process transitioning from \IstyleB to \IstyleA. Notably, stylization results maintain consistent visual quality irrespective of variations in \IstyleA.}
    \vspace{-10pt}
    \label{fig:abl3}
\end{figure}

\section{Additional Experiments and Limitations}

\subsection{Experiments Settings}
The encoding timestep $t_0$ is set to 500, equating to half of the total timesteps. Based on empirical findings, we define the loss function parameters as $\lambda_{re-image}=10$, $\lambda_{re-lpips}=10$, $\lambda_{re-clip}=30$, $\lambda_{cross}=1$, and $\lambda_{in}=0.5$. For the SPN, we assign $\lambda_{SPN}=0.1$. During sampling, $f_{ch}$ is configured to 32, indicating that \zsemstyle conditions up to the 32-resolution feature maps while other blocks are conditioned on \zsemin.

\subsection{Reliance on $I_A^{style}$}
The generation of \IstyleA\ is inherently stochastic, leading to variations in its visual quality. However, the efficacy of our method is not intrinsically tied to the visual fidelity of \IstyleA. As depicted in Figure~\ref{fig:abl3}, despite the varying visual presentations of \IstyleA, our method consistently produces reliable stylization results. The resilience of our approach stems from the stylization trajectory determined in the CLIP space, thereby decoupling it from the aesthetic variations of \IstyleA. To mitigate any misalignment that might arise between \IstyleA and \IstyleB, we adopt a systematic sampling methodology, subsequently auto-selecting the most congruent image, as outlined in \textbf{Training Preparation}.

\subsection{Additional Results}
\begin{figure*}\centering
    \includegraphics[width=\linewidth]{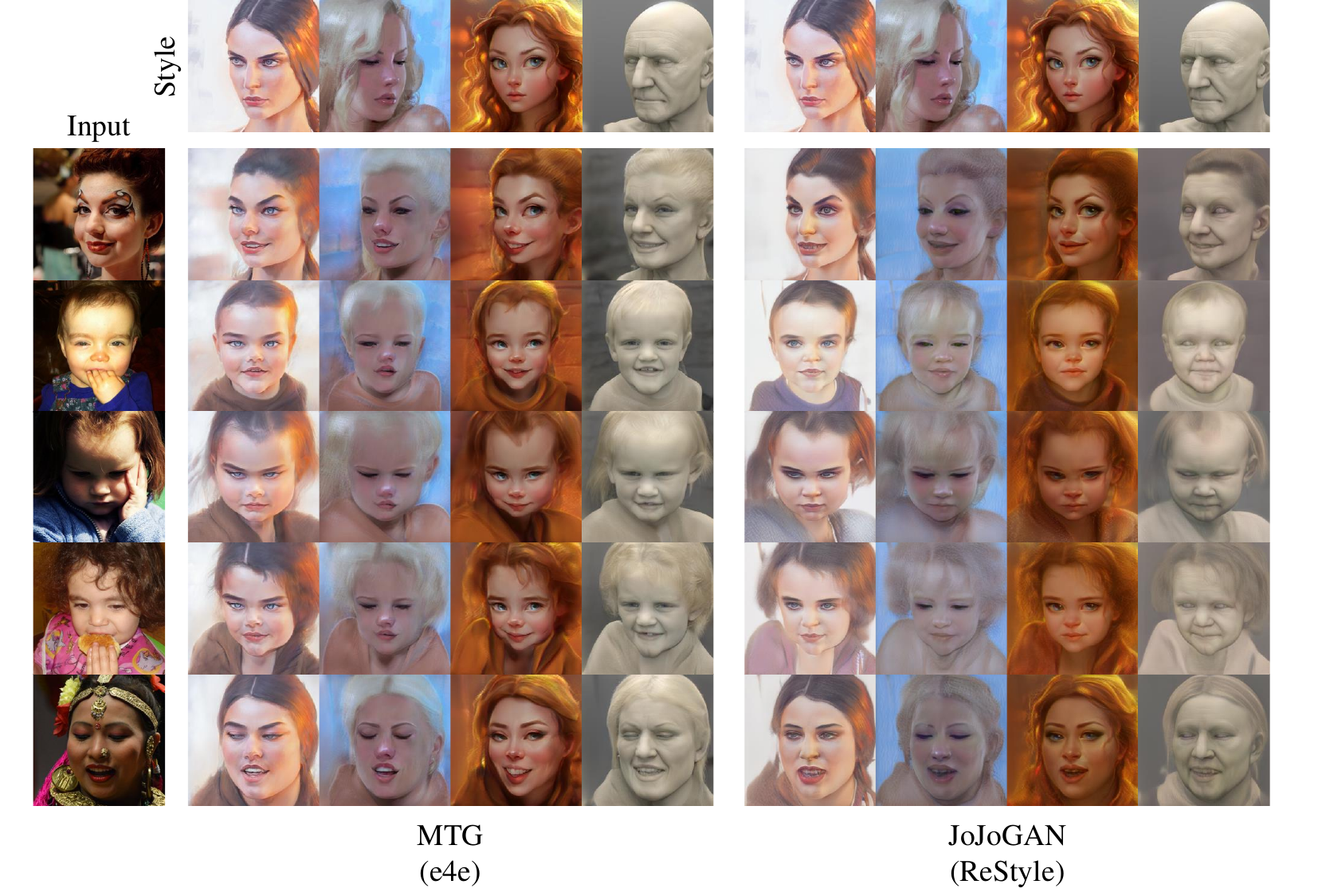}
    \caption{Stylization results of MTG and JoJoGAN.}
    \label{fig:MTG}
\end{figure*}

\begin{figure*}\centering
    \includegraphics[width=\linewidth]{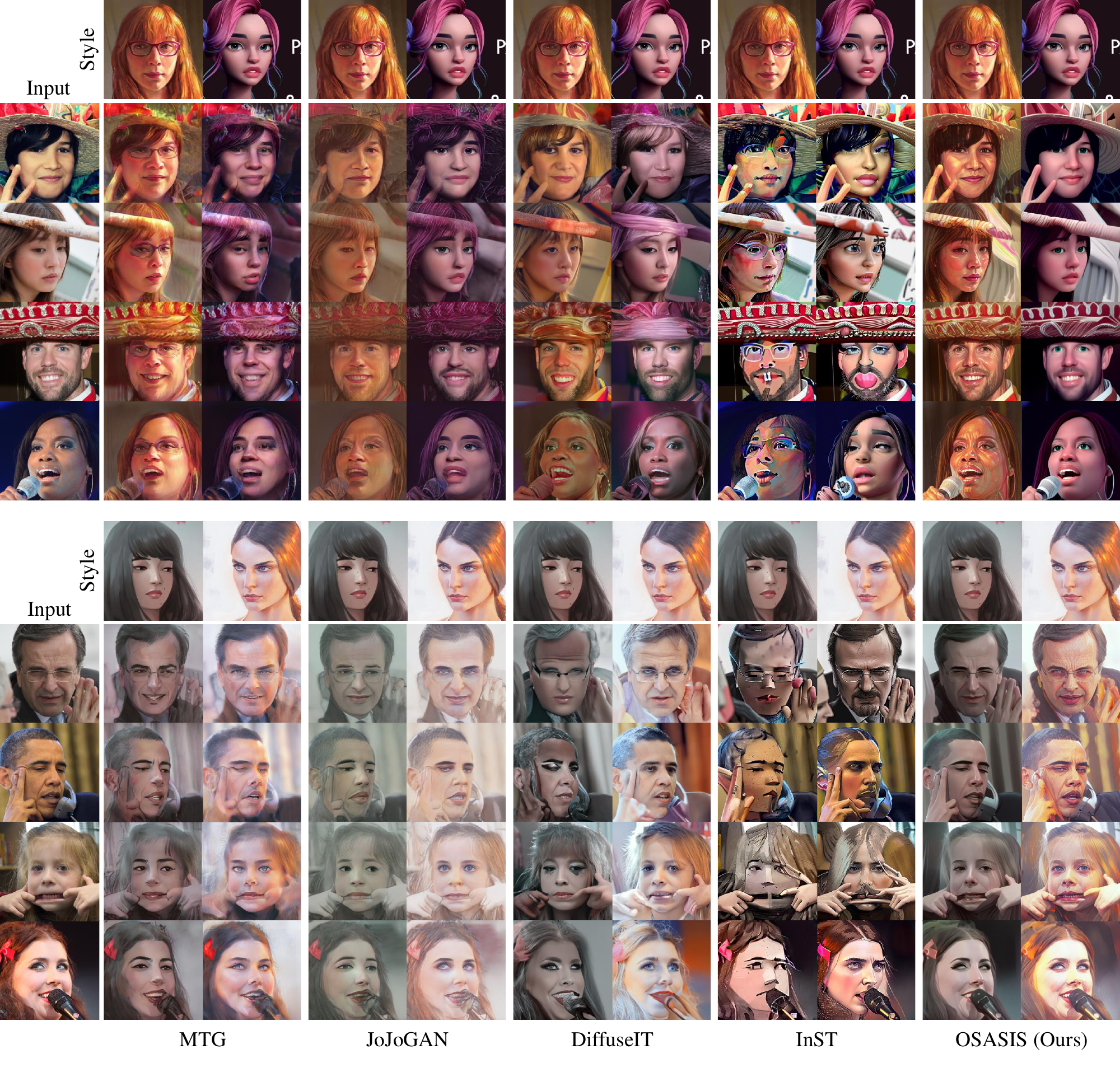}
    \vspace{-20pt}
    \caption{Stylization result of OSASIS and comparison methods.}
    \label{fig:supp_face}
\end{figure*}

\begin{figure*}\centering
    \includegraphics[width=\linewidth]{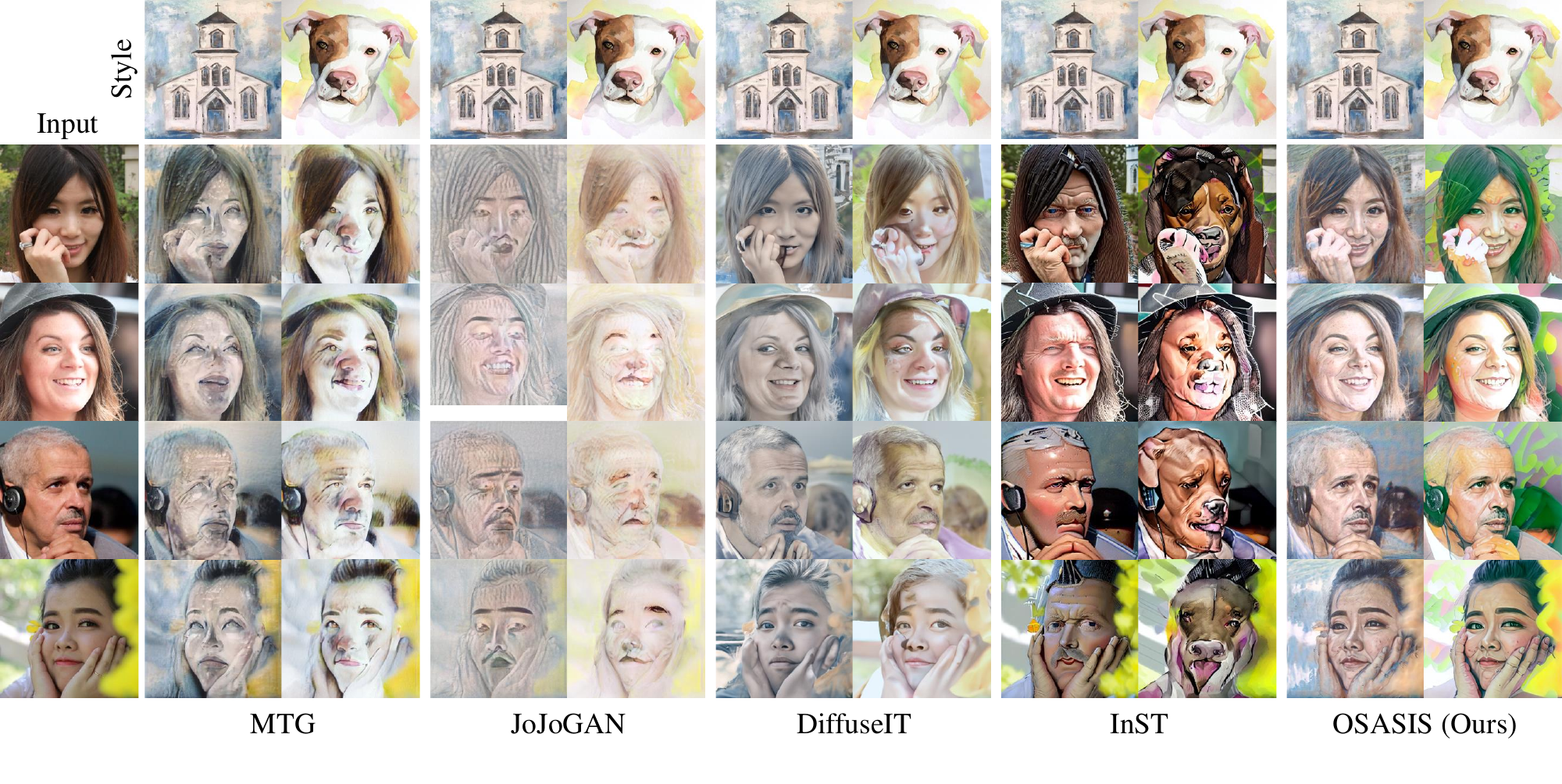}
    \vspace{-20pt}
    \caption{Stylization result of OSASIS and comparison methods with OOD reference images.}
    \label{fig:supp_face_ood}
\end{figure*}

In this section, we present additional stylization results and its comparisons with other stylization methods. Figure~\ref{fig:MTG} shows the stylization outcomes of the original MTG~\cite{zhu2021mind} and JoJoGAN~\cite{chong2022jojogan}, which utilizes e4e~\cite{tov2021designing} and ReStyle~\cite{alaluf2021restyle} respectively for inversion. Compared to HFGI~\cite{wang2022high}, these methods struggle to preserve the structure of the input image leading to a loss of key elements, such as hands and accessories. Additional stylization results on comparison methods are shown in Figure~\ref{fig:supp_face}, where OSASIS outperforms other methods in structural preservation while stylizing. Figure~\ref{fig:supp_face_ood} shows the stylization results of OOD reference images. Table~\ref{table:quant_compare_full} encapsulates a comprehensive quantitative comparison, encompassing both low and high-density image results.

\subsection{User Evaluation results of qualitative samples}
While quantitative assessments provide quality metrics, user studies offer deeper insights into a stylization model's effectiveness. Thus, we included the results of a preference-based user study in Table~\ref{table:user_study}, where participants evaluated 20 stylization outcomes from OSASIS and its baselines against input and style images. The study documented the selection ratio for each method, aiming to discern the model that most effectively harmonizes input structure with stylistic elements. OSASIS emerged as the favored choice in Table~\ref{table:user_study}, underscoring its excellence in meeting human perceptual standards.

\begin{table}
    \centering
    \begin{tabular}{ccccc}
    \hline
    MTG & JoJoGAN & DiffuseIT & InST & OSASIS(Ours) \\
    \hline
    10.0\% & 20.0\% & 0.0\% & 5.0\% & \textbf{65.0\%} \\
    \hline
    \end{tabular}
    \vspace{-10pt}
    \caption{User study of stylized images from OSASIS and baselines}
    \vspace{-15pt}
    \label{table:user_study}
\end{table}

\begin{table*}
    \centering
    \begin{tabular}{c|l|ccc|ccc|ccc}
    \hline
    & \multirow{2}{*}{Methods} & \multicolumn{3}{c|}{ArtFID$\downarrow$} & \multicolumn{3}{c|}{ID Similarity$\uparrow$} & \multicolumn{3}{c}{Structure Similarity$\downarrow$} \\
    && AAHQ & MetFaces & Prev & AAHQ & MetFaces & Prev & AAHQ & MetFaces & Prev \\
    \hline
    \multirow{8}{*}{\shortstack{High\\Density}}
    & MTG & 34.54 & \textbf{34.73} & 35.18 & 0.2026 & 0.2562 & 0.2219 & 0.0403 & 0.0493 & 0.0477 \\
    & MTG+HFGI & 35.03 & 37.19 & 36.23 & 0.3260 & 0.4362 & 0.3688 & 0.0357 & 0.0375 & 0.0370 \\
    & JoJoGAN & 41.93 & 43.99 & 36.95 & 0.3783 & 0.3477 & 0.3280 & 0.0389 & 0.0465 & 0.0441 \\
    & JoJoGAN+HFGI & 41.20 & 43.32 & 39.29 & 0.4927 & 0.4921 & 0.4353 & 0.0327 & 0.0385 & 0.0346 \\
    & DiffuseIT & 44.03 & 52.92 & 46.56 & 0.6922 & 0.7259 & 0.6970 & \textbf{0.0254} & \textbf{0.0252} & \textbf{0.0250} \\
    & InST & \textbf{31.84} & 46.13 & 30.69 & 0.1760 & 0.1864 & 0.1815 & 0.0390 & 0.0326 & 0.0383 \\
    & \textbf{OSASIS(Ours)} & 33.06 & 41.66 & \textbf{30.46} & \textbf{0.7191} & \textbf{0.7520} & \textbf{0.7303} & 0.0367 & 0.0350 & 0.0345 \\
    \hline
     \multirow{8}{*}{\shortstack{Low\\Density}}
    & MTG & 36.19 & \textbf{36.52} & 35.93 & 0.2228 & 0.2516 & 0.2263 & 0.0608 & 0.0557 & 0.0574 \\
    & MTG+HFGI & 36.39 & 38.02 & 37.27 & 0.3730 & 0.4656 & 0.4063 & 0.0386 & 0.0350 & 0.0360 \\
    & JoJoGAN & 43.51 & 45.23 & 38.49 & 0.3763 & 0.3579 & 0.3319 & 0.0589 & 0.00631 & 0.0605 \\
    & JoJoGAN+HFGI & 40.41 & 44.74 & 41.09 & 0.5145 & 0.5207 & 0.4743 & 0.0411 & 0.0454 & 0.0403 \\
    & DiffuseIT & 44.93 & 53.35 & 48.18 & \textbf{0.6992} & 0.7158 & 0.6994 & \textbf{0.0309} & 0.0300 & \textbf{0.0310} \\
    & InST & 38.16 & 50.33 & 35.86 & 0.2253 & 0.2188 & 0.2238 & 0.0492 & 0.0443 & 0.0488 \\
    & \textbf{OSASIS(Ours)} & \textbf{34.89} & 43.20 & \textbf{33.20} & 0.6825 & \textbf{0.7323} & \textbf{0.7029} & 0.0361 & \textbf{0.0295} & 0.0391 \\
    \hline
    \end{tabular}
    \caption{Quantitative comparison.}
    \vspace{-10pt}
    \label{table:quant_compare_full}
\end{table*}



\end{document}